\documentclass{article}
\PassOptionsToPackage{numbers, compress}{natbib}
 \usepackage[preprint]{neurips_2026}

\usepackage[utf8]{inputenc} 
\usepackage[T1]{fontenc}    
\usepackage[hidelinks]{hyperref}       
\usepackage{url}            
\usepackage{booktabs}       
\usepackage{amsfonts}       
\usepackage{nicefrac}       
\usepackage{microtype}      
\usepackage{xcolor}         
\usepackage{siunitx}        
\usepackage{tcolorbox}
\usepackage{subcaption}
\usepackage{bm} 
\usepackage{acronym}
\usepackage{multirow}
\usepackage{amsmath}
\usepackage{amssymb,amsthm}

\newtheorem{lemma}{Lemma} 
 
\newtheorem{remark}{Remark}

\usepackage[textsize=scriptsize]{todonotes}

\usepackage{array}         
\usepackage{adjustbox}

\usepackage{xspace} 
\usepackage{threeparttable} 
\usepackage{tabularx}
\usepackage{enumitem}
\usepackage{algorithmic}

\usepackage[ruled,vlined,linesnumbered]{algorithm2e}

\usepackage{empheq}
\usepackage{color,soul}

\usepackage[table]{xcolor}

\usepackage{wrapfig}
\usepackage{etoc}

\newcommand{\voxceleb}{VoxCeleb\xspace}
\newcommand{\cnceleb}{CNCeleb\xspace}
\definecolor{pastelgreen}{RGB}{208,239,129}
\definecolor{lavender}{RGB}{222,210,234}
\definecolor{peach}{RGB}{255,198,142}
\newcommand{\graycell}[1]{\cellcolor{gray!20}#1}

\title{Adaptive Regularization for Sparsity Control in Bregman-Based Optimizers}
  \author{
  Ahmad Aloradi\textsuperscript{1,2}\quad
  Tim Roith\textsuperscript{3,4}\quad
  Emanu\"el A.~P.~Habets\textsuperscript{2}\quad
  Daniel Tenbrinck\textsuperscript{1}\\[2pt]
  \textsuperscript{1}Department of Data Science, FAU Erlangen-N\"urnberg, Erlangen, Germany\\
  \textsuperscript{2}International Audiolabs, FAU Erlangen-N\"urnberg, Erlangen, Germany\\
  \textsuperscript{3}School of Computation, Information and Technology, Technical University of Munich\\
  \textsuperscript{4}Munich Center for Machine Learning (MCML)\\[3pt]
  \texttt{\{ahmad.aloradi, daniel.tenbrinck\}@fau.de}\quad
  \texttt{tim.roith@tum.de}\quad\\
  \texttt{emanuel.habets@audiolabs-erlangen.de}
  }

\DeclareMathOperator*{\argmin}{arg\, min}
\DeclareMathOperator*{\argmax}{arg\, max}
\newcommand{\loss}{\mathcal{L}}
\newcommand{\func}{\mathcal{\phi}}
\newcommand{\R}{\mathbb{R}}
\newcommand{\EN}{\operatorname{EN}}
\DeclareMathOperator{\prox}{prox}

\newcommand{\dummy}{\mathord{\color{black!33}\bullet}}

\usepackage{tikz}
\usetikzlibrary{tikzmark,calc,decorations.pathreplacing}
\begin{document}

\newacro{AAM}{Additive Angular Margin Softmax}
\newacro{AS-norm}{adaptive score normalization}
\newacro{PLDA}{probabilistic linear discriminant analysis}
\newacro{BN}{bottleneck feature}
\newacro{RTF}{real time factor}
\newacro{DNN}{deep neural network}
\newacro{EER}{equal error rate}
\newacro{minDCF}{Minimum detection Cost function}
\newacro{MSE}{mean-squared error}
\newacro{NSF}{neural source-filter}
\newacro{AM}{acoustic model}
\newacro{VPC}{VoicePrivacy Challenge}
\newacro{VPAC}{VoicePrivacy Attacker Challenge}
\newacro{SV}{speaker verification}
\newacro{ASV}{automatic speaker verification}
\newacro{ASR}{automatic speech recognition}
\newacro{TDNN}{time delay neural network}
\newacro{FPE}{Fine Pitch Error}
\newacro{GPE}{Gross Pitch Error}
\newacro{MFCC}{Mel Frequency Cepstral Coefficients}
\newacro{CNN}{convolutional neural network}

\newacro{OOD}{out-of-distribution}
\newacro{ID}{in-distribution}

\newacro{PC}{poorly-conditioned}
\newacro{RC}{reasonably-conditioned}

\newacro{TSTP}{temporal statistics temporal pooling}
\newacro{ASTP}{attentive statistics temporal pooling}

\newacro{GD}{gradient descent}
\newacro{MD}{mirror descent}
\newacro{LinBreg}{linearized Bregman}
\newacro{DA}{Nesterov's dual averaging}

\newacro{LTH}{lottery ticket hypothesis}
\newacro{SET}{sparse evolutionary training}
\newacro{DST}{dynamic sparse training}

\maketitle

\begin{abstract}
Sparse training reduces the memory and computational costs of deep neural networks. However, sparse optimization methods, e.g., those adding an $\ell_1$ penalty, often control sparsity only indirectly through a regularization parameter $\lambda$, whose mapping to the final sparsity rate is non-trivial. In our experiments, we found this parameter sensitivity to be particularly pronounced for Bregman-based optimizers. Specifically, the two variants \texttt{LinBreg} and \texttt{AdaBreg} reach the same sparsity at $\lambda$ values that differ by up to two orders of magnitude, requiring expensive trial-and-error sweeps to achieve a user-specified sparsity. To address this, we propose an adaptive regularization scheme that updates $\lambda$ based on the difference between the model's current sparsity and the target sparsity. We analyze the resulting algorithm and evaluate it on automatic speaker verification with ECAPA-TDNN and ResNet34 on VoxCeleb and CNCeleb. The proposed method reliably achieves sparsity targets ranging between 75\% and 99\%. It also converges faster than the oracle-tuned non-adaptive baseline during early training and matches or surpasses its final performance in equal error rate. We further show that the adaptive scheme inherits key properties from its non-adaptive counterpart, including improved out-of-distribution robustness over the dense baselines.
\end{abstract}

\section{Introduction}

\subsection{Background}
The success of \acp{DNN} across a wide range of applications 
\cite{openai2024gpt4technicalreport, jumper2021alpha_fold} has been accompanied by an exponential increase in scale, both in the number of parameters and in the compute required for training \cite{villalobos2022compute}. Consequently, both memory requirements and the demand for specialized hardware have grown dramatically. This, in turn, has led to growing concern regarding the environmental cost of scaling \acp{DNN} \cite{dhar2020carbon, luccioni2023estimating, strubell2019energy}.

In response, a significant body of research is devoted to developing more efficient models. Learning sparse models, or \emph{sparsification}, is one such technique for improving efficiency. A comprehensive review is provided in~\cite{hoefler2021sparsity}. In the context of weight sparsity, we broadly divide the methods into three main categories: \emph{pruning}, \emph{\ac{DST}}, and \emph{sparse optimization}.

In classical pruning, a dense network is trained, after which parameters with the smallest magnitude are permanently removed \cite{LeCun_optimal_brain_damage, han2015learning}. Pruning methods vary widely, encompassing unstructured versus structured sparsity, as well as one-shot versus iterative pruning schedules \cite{janusz2024oneshot}. 
In contrast to pruning, \ac{DST} typically follows a \emph{train-prune-grow} cycle to dynamically adjust which weights are active. For instance, \ac{SET} \cite{mocanu2018scalable} utilizes magnitude-based pruning and regrows connections at random while keeping sparsity fixed. In \cite{evci2020rigging}, the random weight regrowth was replaced by a gradient-informed criterion, whereas \cite{dettmers2019sparsenetworksscratchfaster} uses momentum. Although most \ac{DST} methods optimize the weight objective with SGD, they often rely on heuristics for the mask objective \cite{ji_advancing_dst}.

Sparse optimization takes a different approach to sparsity by incorporating sparsity-inducing penalties into the training objective. A common approach is to add an explicit $\ell_1$ regularization term \cite{tibshirani1996regression}, which serves as a convex relaxation of the exact $\ell_0$ penalty. A differentiable $\ell_0$ relaxation via non-negative stochastic gates to penalize non-zero weights was presented in \cite{louizos2018learning}.
Other approaches use \emph{implicit} regularization, rather than an explicit penalty term, which can be achieved using \ac{MD} \cite{yudin}. A series of works \cite{Azizan, bregman_leon, huang2016split,lunk2026multilevel, lifted_bregman} leverage \ac{MD} and \emph{linearized Bregman iterations} to induce sparsity without explicit regularization.

\begin{wrapfigure}[15]{R}{0.45\textwidth}
    \centering
    \vspace{-9pt}
    \begin{subfigure}[tb]{0.4\textwidth}
    \centering
    \includegraphics[width=\linewidth]{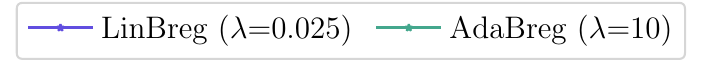}
    \end{subfigure}\\
    \begin{subfigure}[tb]{0.4\textwidth}
    \centering
    \includegraphics[width=\linewidth]{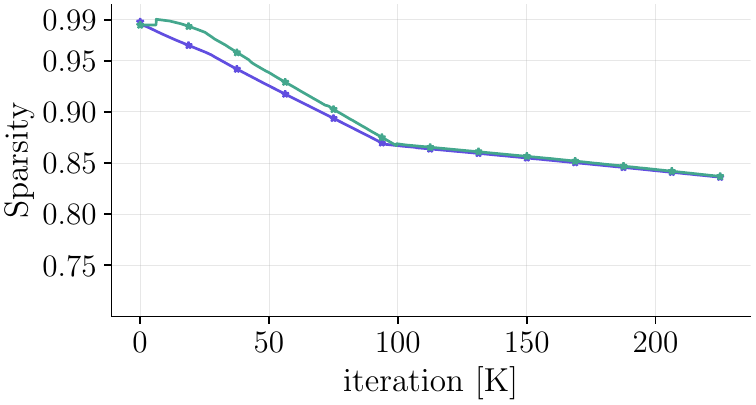}
    \end{subfigure}
    %
    \caption{Sparsity profiles of the two Bregman optimizers: \texttt{LinBreg} and \texttt{AdaBreg}. The sparsity changes during training, and the same final sparsity can be obtained using $\lambda$ values that differ by a factor of $400$.}
    \label{fig:intro_bregman}
\end{wrapfigure}

\subsection{Sparsity control in sparse optimization frameworks}
A key challenge in sparse optimization is controlling the sparsity rate of the trained model. Sparse optimization methods typically control sparsity indirectly via a regularization parameter, often denoted by $\lambda$. Mapping a given $\lambda$ to a predetermined sparsity rate is highly non-trivial. Figure~\ref{fig:intro_bregman} demonstrates this for the \ac{MD} setup in \cite{bregman_leon}. 

Incorporating sparsity constraints in sparse optimization has been studied in various settings. \citet{kim2016deep} adjusted per-layer $\ell_1$ coefficient during training based on the gap between the current and target non-zero ratio in each layer. Their work assumes a fully-connected architecture with predefined per-layer targets. 
\citet{shen2024sparse} characterized the sparsity of $\ell_1$-regularized networks as a function of~$\lambda$ and developed an iterative selection algorithm to solve for the parameter that meets the sparsity constraint. However, each iteration requires re-training, rendering the procedure expensive at scale.
In \cite{HyperSparse2023}, a multi-stage training approach is employed, where $\lambda$ is adapted via a prescribed schedule over epochs during the finetuning stage. The general approach in~\cite{mackay2018selftuning} casts hyperparameter adaptation as a bilevel optimization problem. However, they investigated $\ell_2$ decay and dropout rather than sparsity-inducing penalties, and they incur the overhead of maintaining a best-response approximation.


\subsection{Contributions}
This paper enhances the Bregman learning framework of \citet{bregman_leon} by introducing an automatic $\lambda$ adaptation for controlling the sparsity rate. 
Similar to \cite{HyperSparse2023}, we adapt $\lambda$ during training based on the difference from the target sparsity rather than a fixed schedule. Unlike \cite{HyperSparse2023,shen2024sparse}, our scheme uses an \ac{MD} setup rather than an $\ell_1$-regularized loss. Our main contributions are summarized as follows:
\begin{enumerate}[label=(\Roman*), leftmargin=*]
    \item We extend the framework in~\cite{bregman_leon} by an adaptive regularisation parameter $\lambda$ to train neural networks with a specific target sparsity. We empirically show that models trained with our automatic scheme converge to solutions that perform on par with, or even better than, those trained with an oracle $\lambda$ that leads to a similar target sparsity.
    \item We investigate the robustness of sparse models, produced by pruning and the Bregman approach, against their dense counterparts on \ac{OOD} data. We observe that sparsity tends to induce robustness, and that adapting $\lambda$ preserves this robustness.
    \item We provide a detailed analysis of the proposed framework and demonstrate its scalability using \ac{ASV} as an application.
    \item We identify sub-optimal sparsity allocation as a source of failure in Bregman optimizers at extremely high sparsity.
    \end{enumerate}
    









\section{Sparse Training via Adaptive Regularization}

The goal in this work is the training of neural networks under a sparsity constraint. Particularly, for some supervised dataset $\mathcal{T}$ we minimize the loss function $\loss(\theta) := \sum_{(x,y)\in\mathcal{T}} \ell\bigl(F(x;\theta),\,y\bigr)$, where $\theta\in\Theta\subset\R^d$ are parameters of a neural network $F(\dummy;\theta)$ and $\ell(\dummy,\dummy)$ denote the per-sample loss. To compute a sparse solution, we prescribe a target number of non-zero parameters $\mathsf{a}^*\in\mathbb{N}$ and solve
\begin{align}\label{eq:spopt}
\min_{\theta: |\theta|_0 \leq \mathsf{a}^*} \loss(\theta),
\end{align}
where $|\dummy|_0$ denotes the $\ell_0$-functional on $\R^d$. This is connected to the sparsity, which we define as
\begin{align*}
\mathsf{s}^* := \frac{d - \mathsf{a}^*}{d},\qquad \mathsf{s}(\theta) := \frac{d - |\theta|_0}{d}.
\end{align*}

\begin{remark}
%
In the context of neural networks, the so-called \emph{Lottery Ticket Hypothesis} \cite{frankle2018the} suggests the existence of sparse sub-networks that match the performance of their dense counterparts. In our notation, this translates to the existence of $\mathsf{a}^* < d$ such that $\min_{\theta: |\theta|_0 \leq \mathsf{a}^*} \loss(\theta) = \min_{\theta\in\Theta} \loss(\theta)$.
\end{remark}
In order to solve \eqref{eq:spopt}, we employ an iterative scheme based on the inverse scale space approaches introduced in \cite{bregman_leon,osher2005iterative}. That is, we start with an initialization $\theta^{(0)}$ with very few active components and in each iteration $k$ desire the following properties:
\begin{tikzpicture}[remember picture, overlay]
\node (rightenum) at (.075\textwidth,0) {};
\draw [decorate, decoration={brace}, thick] ($({pic cs:top1} -| rightenum) + (0, 1em)$) -- ({pic cs:bot1} -| rightenum) node [midway, right] {\color{blue!50}\quad Bregman learning framework \cite{bregman_leon}};
\draw [decorate, decoration={brace}, thick] ($({pic cs:top2} -| rightenum) + (0, 1em)$) -- ({pic cs:bot2} -| rightenum) node [midway, right] {\color{green!60!black}\quad Our contribution};
\end{tikzpicture}
\begin{enumerate}
\item[(I1)]\tikzmark{top1} $\loss(\theta^{(k+1)}) < \loss(\theta^{(k)})$: loss decay,
\item[(I2)]\tikzmark{bot1} $|\theta^{(k+1)}|_0\approx |\theta^{(k)}|_0 $: sparsity preservation,
\item[(I3)]\tikzmark{top2}\tikzmark{bot2} $|\theta^{(k+1)}|_0 \approx \mathsf{a}^*$: sparsity constraint.
\end{enumerate}
We address (I1) and (I2) by defining the next iterate to be the solution of the following problem,
\begin{align}\label{eq:it}
\theta^{(k+1)} = \argmin_{\theta\in\Theta}\,\langle \nabla \loss(\theta^{(k)}), \theta - \theta^{(k)}\rangle + \frac{1}{\tau}\mathcal{R}(\theta, \theta^{(k)}),
\end{align}
where we linearized $\loss$ around the previous iterate, i.e., we replace $\loss(\theta)$ by $\langle \nabla \loss(\theta^{(k)}), \theta\rangle$. The functional $\mathcal{R}:\Theta^2\to\R\cup\{+\infty\}$ promotes (I2), where its strength is weighted by the parameter $\tau>0$. 
While this can be handled by the framework developed in \cite{bregman_leon}, it does not yield any control over the actual sparsity of $\theta^{(k+1)}$, only over its increment compared to $\theta^{(k)}$. The main contribution of this paper is to address (I3), i.e., keeping the iterates within a desired sparsity regime. This can be achieved through a suitable adaptation strategy for the functional $\mathcal{R}$, as described in the following.
\subsection{Mirror Descent and linearized Bregman iterations}
The core element of \eqref{eq:it} is the functional $\mathcal{R}$. We note that, for example $\mathcal{R}(\theta,\tilde{\theta})=\nicefrac{1}{2}|\theta - \tilde{\theta}|_2^2$ would yield a standard gradient descent scheme. 
A popular choice is the \emph{Bregman divergence}\footnote{We refer to \cite{Bauschke2017,rockafellar1998variational} for an introduction to convex analysis and related concepts.}
\begin{align*}
D_{\func}^p(\theta, \tilde{\theta}) := \func(\theta) - \func(\tilde{\theta}) 
- \langle p, \theta - \tilde{\theta} \rangle,\qquad p\in\partial\func(\tilde{\theta}),
\end{align*}
where $\partial\func$ denotes the subdifferential of the convex functional $\func:\Theta\to\R$. Setting $\mathcal{R}(\theta,\theta^{(k)})=D^{p^{(k)}}_\func(\theta, \theta^{(k)})$ with $p^{(k)}\in\partial\func(\theta^{(k)})$ in \eqref{eq:it} leads to a family of methods known in the literature as \emph{mirror descent}, which we can write as
\begin{align}\label{eq:md}
\theta^{(k+1)} = \argmin_{\theta\in\R^d}\left\{\langle \theta - \theta^{(k)}, \nabla \loss(\theta^{(k)})\rangle + \frac{1}{\tau}D_{\func}^{p^{(k)}}(\theta, \theta^{(k)})\right\} = 
\nabla\func^*\left(p^{(k)} - \tau \, \nabla\loss(\theta^{(k)})\right),
\end{align}
where $\nabla\func^* = (\partial \func)^{-1}$ is the gradient of the convex conjugate $\func^*$. Applying $\partial\phi$ to both sides in \eqref{eq:md}, we observe that setting $p^{(k+1)} = p^{(k)} - \tau \, \nabla\loss(\theta^{(k)})\in \partial\func(\theta^{(k+1)})$ yields a valid sequence of subgradients. This is then known as \emph{lazy mirror descent}, 
\ac{DA} \cite{nesterov2009primal}, or \ac{LinBreg} iterations \cite{yin2008bregman,cai2009linearized}. Furthermore, by replacing the simple gradient step with a momentum or Adam version \cite{kingma2015adam}, we obtain \emph{AdaBreg} as introduced in \cite{bregman_leon}. Regarding the choice of $\phi$, we adopt the standard approach from compressed sensing (cf. \cite{candes2006stable}) and replace the $\ell_0$-functional by the sparsity-promoting and convex $\ell_1$-norm. To ensure strong convexity, we use the elastic net functional
\begin{align*}
\EN_\lambda(\theta) := \frac{1}{2}|\theta|_2^2 + \lambda |\theta|_1
\quad\Rightarrow\quad 
\nabla \EN^*_\lambda (p) = 
\prox_{\lambda |\dummy|_1}(p) = 
(\operatorname{sign}(p_i) \max\{|p_i| - \lambda, 0\})_{i=1}^d,
\end{align*}
where $\prox_\mathcal{R}(p):=\argmin_{w} \frac{1}{2} |p-w|_2^2 + \mathcal{R}(w)$ denotes the proximal operator. 

\subsection{Promoting the target sparsity via parameter adaptation}

In the classical setting of \ac{LinBreg} one considers a loss function $\loss(\theta) = \ell(A \, \theta, y)$ for some linear operator $A$ and function $\ell$.
One can show that with suitable step sizes the iterates converge to the closest minimizer in Bregman distance, that is $\theta^{(k)} \xrightarrow[]{k\to\infty} \argmin_{\theta:\loss(\theta)=0} D^{p^{(0)}}_\func(\theta,\theta^{(0)}),$
see \cite{gunasekar2018characterizing,zhang2010bregmanized,yin2008bregman}. While this illustrates the algorithm's implicit bias, we cannot expect a similar result to hold in our case. 
In general, the dependence on the parameter $\lambda$ of the limit of $\theta^{(k)}$ as $k\to\infty$ is hard to characterize, and there is no explicit selection rule for $\lambda$ available. Instead, we adapt it iteratively to ensure that (I3) is fulfilled. Namely, for a sequence of parameters $\lambda^{(k)}\geq 0$, \eqref{eq:md} is modified to
\begin{align}\label{eq:mdup}
p^{(k+1)} = p^{(k)} - \tau \, \nabla\loss\left(\theta^{(k)}\right),\qquad
\theta^{(k+1)} = 
\nabla\EN_{\lambda^{(k)}}^*\left(p^{(k+1)}\right).
\end{align}
The parameter $\lambda_k$ is chosen based on the current sparsity, namely increased if $|\theta^{(k)}|_0 > \mathsf{a}^*$ and decreased otherwise. This modification introduces a tension in the setup of \eqref{eq:md}, in the sense that now  $p^{(k+1)}\in \partial \EN_{\lambda^{(k)}}(\theta^{(k+1)})$ but possibly $p^{(k+1)}\notin\partial\EN_{\lambda^{(k+1)}}(\theta^{(k+1)})$. Nevertheless, due to the specific form of our parameter adaptation, we directly obtain the following decay up to an error that can be controlled by the magnitude of the change in $\lambda^{(k)}$. The proof uses standard arguments and is included in Appendix~\ref{app:proof} for completeness. We refer to \cite{benfenati2013inexact} for a related convergence analysis of inexact Bregman methods and leave the stochastic setting as in \cite{bregman_leon} for future work.
\begin{lemma}\label{lem:lossdecay}
Assuming that $\loss$ is $L$-smooth, for the iterates in \eqref{eq:mdup} with parameters $\lambda^{(k)}\geq 0$ we obtain
\begin{align*}
\loss(\theta^{(k+1)}) + \left(\frac{1}{\tau} - \frac{L}{2}\right) |\theta^{(k+1)} - \theta^{(k)}|_2^2 
+
\frac{\lambda^{(k)} - \lambda^{(k-1)}}{\tau}( |\theta^{(k+1)}|_1 - |\theta^{(k)}|_1)
\leq \loss(\theta^{(k)}).
\end{align*}
\end{lemma}
Alternatively, one could perform a subgradient correction scheme, to reintroduce the exact subgradient property. However, empirically, we found that such schemes performed worse than the simple update in \eqref{eq:mdup}, which is discussed in Appendix~\ref{app:corr}.

We briefly mention related concepts of adaptive parameter choices in mirror descent. The dual averaging formulation in \cite{nesterov2009primal} allows for an adaptive parameter that rescales the whole functional $\phi$. We comment on the conceptual difference in Appendix~\ref{app:proxrescale}. Furthermore, \cite{benning2017choose} applies an adaptive technique, where the parameter $\lambda$ is linked with the step size $\tau$. Beyond that, mirror descent with adaptive functional $\phi$ is also known as \emph{variable Bregman majorization-minimization} \cite{adly2026variable,martin2025variable}, where the functional $\phi$ is often assumed to be differentiable.

\subsection{Proposed parameter update strategy}
\label{sec:lambda_update}

To control the regularization parameter $\lambda^{(k)}$, we define the sparsity defect as $\epsilon^{(k)} := \mathsf{s}^{\ast} -  \mathsf{s}(\theta^{(k)})$. 
To relax the objective $\mathsf{s}(\theta)=\mathsf{s}^\ast$, we allow for a small tolerance $\zeta$ such that
\begin{equation}
    \label{eq:sparsity_tolerance}
    \lvert \mathsf{s}(\theta) - \mathsf{s}^\ast\rvert \: \leq \: \zeta, \qquad 1\gg\zeta>0.
\end{equation}
We propose an adaptive scheme that updates the regularization parameter $\lambda^{(k)}$ only every $f \in \mathbb{N}$ steps.
In particular, we use the following update strategy:
\begin{equation}\label{eq:lambda-update}
\lambda^{(k+1)} = 
\begin{cases} 
\lambda^{(k)} \bigl(1 + \alpha\,\lvert \epsilon^{(k)}\rvert\bigr)^{\,\operatorname{sgn}(\epsilon^{(k)})} & \text{if } k \bmod f = 0,\\
\lambda^{(k)}     & \text{else,}
\end{cases}
\end{equation}
where $\operatorname{sgn}$ is the sign function and $\alpha>0$ acts as an acceleration factor.
Smaller values of $\alpha$ ensure controlled updates, whereas larger values result in a fast convergence towards the target sparsity $\mathsf{s}^*$. To ensure stability as the iterates approach the tolerance band in \eqref{eq:sparsity_tolerance}, we introduce a \textit{damping} mechanism.
We define a threshold $\zeta_d$ (where typically $\zeta_d \approx \zeta$) and adapt both the update frequency $f$ and the acceleration factor $\alpha$ when the error is small via: $(f^{(k+1)}, \alpha^{(k+1)}) = (\gamma_f f^{(k)} , \, \gamma_\alpha \alpha^{(k)})$ if $\bigl|\epsilon^{(k)} \bigr| \leq \zeta_d$,
with $\gamma_f \in \mathbb{N}_{\geq 2}$ and $\gamma_\alpha \in \mathbb{R}^+$.
This strategy reduces the adaptation frequency and step size once the target sparsity is nearly reached, effectively preventing high-frequency oscillations of the sparsity rate around $\mathsf{s}^*$.
The latter can occur when the distribution of equivalent model weights is concentrated densely around the current value of $\lambda$, such that small updates of $\lambda$ can lead to large changes in $\epsilon^{(k)}$.
Although one could stop the adaptation entirely when the tolerance \eqref{eq:sparsity_tolerance} is satisfied, we maintain the update rule in our experiments to evaluate the long-term stability and robustness of our proposed adaptive regularization scheme. 
%
%
%
Algorithm~\ref{algo:lambda_adaptation} summarizes this adaptation scheme together with the update from \eqref{eq:mdup}. An analysis of the algorithm's properties is provided in Appendix~\ref{sec:proposed_properties}.

\begin{algorithm}[t]
\DontPrintSemicolon
\small
\caption{Bregman iterations with $\lambda$ adaptation strategy}
\label{algo:lambda_adaptation}
\KwIn{Target sparsity $\mathsf{s}^{\ast}$, adaptation frequency $\; f,\;$ adaptation control $\alpha,\;$ learning rate $\tau$}
\textbf{Initialize:} $\lambda^{(0)},\;$ $p^{(0)},\; \theta^{(0)},\; \epsilon^{(0)} \leftarrow \mathsf{s}^\ast {-} \mathsf{s}\bigl(\theta^{(0)}\bigr)$\\
\For{$k = 0, 1, 2, \ldots$}{
    Sample mini-batch $\mathcal{B} \sim \mathcal{T}$\\
    $p \leftarrow p - \tau \nabla\mathcal{L}_{\mathcal{B}}(\theta)$ \tcp*{subgradient step}
    $\theta \leftarrow \operatorname{prox}_{\lambda \mathcal{R}}(p)$ \tcp*{Proximal step}
    $\epsilon^{(k)} = \mathsf{s}^\ast -  \mathsf{s}(\theta^{(k)})$\\
    \If{$k \bmod f = 0$}{
        $\lambda \leftarrow \lambda (1 + \alpha|\epsilon^{(k)}|)^{\operatorname{sgn}(\epsilon^{(k)})}$\tcp*{Update ElasticNet parameter}
    }
    \If{$|\epsilon^{(k)}| < \zeta_d$}{
        $f \leftarrow \gamma_f f,\; \alpha \leftarrow \gamma_\alpha \alpha$ \tcp*{Reduce size and frequency of the updates}
    }
}
\end{algorithm}

\section{Experimental Setup}
\label{sec:exp_setup}

\subsection{Datasets and preprocessing}
\label{subsec:datasets}

\begin{table}[tb]
    \centering
    \footnotesize
    \caption{\voxceleb and \cnceleb datasets details} 
    \label{tab:datasets}
    \renewcommand{\arraystretch}{0.4}
    \setlength{\tabcolsep}{12pt}
    \begin{tabular}{@{} l l c r r r @{}}
        \toprule
        \textbf{Corpus} & \textbf{Split} & \textbf{Subset} & \textbf{\# Speakers} & \textbf{\# Utterances} & \textbf{\# Trials} \\
        \midrule
        \multirow{4}{*}{\voxceleb}   & Dev                   & \voxceleb~2 dev set & \num{5994} & \num{1128246} & -- \\
        \cmidrule{2-6}
                                         & \multirow{3}{*}{Test} & \voxceleb1-O                & \num{40}   & \num{4708}   & \num{37611} \\
                                         &                       & \voxceleb1-E                & \num{1251} & \num{145160} & \num{579818} \\
                                         &                       & \voxceleb1-H                & \num{1190} & \num{137924} & \num{550894} \\
        \midrule
        \multirow{2}{*}{\cnceleb}    & Dev                   & \cnceleb2 + 1 dev sets & \num{2793} & \num{533929} & -- \\
        \cmidrule{2-6}
                                         & Test                  & \cnceleb-E                & \num{200}  & \num{17777}  & \num{3484292} \\
        \bottomrule
    \end{tabular}
\end{table}
\vspace{-2pt}

We used two popular \ac{ASV} benchmarks: \voxceleb~\cite{voxceleb}, which is multilingual, and \cnceleb~\cite{cnceleb}, which is Chinese and spans $11$ genres. Both are sampled at $16$~kHz and consist of real-world, non-curated speech from open-source multimedia. Each comprises two subsets (\voxceleb~1, \voxceleb~2 and \cnceleb~1, \cnceleb~2); Table~\ref{tab:datasets} summarizes their key features. We trained on \voxceleb~2 (dev) and evaluated on all three \voxceleb~1 test sets. We combined \cnceleb~2 with the \cnceleb~1 dev split to form our development set (\cnceleb-D) and evaluated on the \cnceleb~1 test set (\cnceleb-E). In both benchmarks, dev and test speakers do not overlap. Unlike \voxceleb, \cnceleb-E uses multi-session enrollment. We computed one embedding per utterance and aggregated them by a length-weighted mean to form the speaker embedding.

Audio was segmented into $3.0$-second non-overlapping chunks, mean-centered, normalized to $-20$~dB root-mean-square level, and peak-clipped at $1.0$. For \cnceleb, where many recordings are shorter than $2$~s, we first concatenated same-speaker same-genre clips to lengths above $5$~s before segmentation. We then extracted $80$ log Mel-filterbank features with a $512$-point FFT, $25$~ms window, and $10$~ms hop, followed by utterance-level mean normalization. \emph{No data augmentation} is used\footnote{Extensive augmentation is standard in state-of-the-art \ac{ASV} recipes on these corpora. We omitted it to enable a controlled robustness analysis.}.

\begin{wraptable}[11]{R}{0.5\textwidth}
    \centering
    \footnotesize 
    \vspace{-3pt}
    \caption{Preprocessing and training configurations.}
    \label{tab:training-config}
    \renewcommand{\arraystretch}{0.7}
    \setlength{\tabcolsep}{1pt}
    \begin{tabular}{@{}lcc@{}}
        \toprule
        \multirow{2}{*}{\textbf{Hyperparameter}} & \multicolumn{2}{c}{\textbf{Dataset}}  \\
        \cmidrule(ll){2-3} 
        & \voxceleb & \cnceleb \\
        \midrule
        \# Epochs                                & $20$ & $40$ \\
        \# Speech segments           & \num{2161041} & \num{1154912}\\
        Train / validation split                 & $95\% / 5\%$ &  $90\% / 10\%$\\
        \cmidrule(rl){1-3} 
        \acs{AAM}-Softmax Loss                    & \multicolumn{2}{c}{scale ${=} 32$, margin ${=} 0 \to 0.2$} \\
        Batch size                               & \multicolumn{2}{c}{256 (ECAPA-TDNN), 128 (ResNet34)} \\
        \bottomrule
    \end{tabular}
\end{wraptable}

\subsection{Implementation details}
\label{subsec:implementation}

We used ECAPA-TDNN~\cite{desplanques_ecapa-tdnn_2020} (\texttt{C1024} variant, $192$-dim embeddings, attentive statistics pooling) and ResNet34 ($256$-dim embeddings, temporal statistics pooling), both from the WeSpeaker toolkit~\cite{wang2023wespeaker}.
The ECAPA-TDNN architecture contains \num{14.7}M parameters, with the classifier adding \num{0.72}M parameters on \cnceleb and \num{1.4}M on \voxceleb. The ResNet34 architecture consists of \num{6.6}M parameters, with its classifier adding \num{0.72}M on \cnceleb and \num{1.8}M on \voxceleb.
All models were trained from scratch with the Additive Angular Margin (AAM) Softmax loss~\cite{aam_softmax} on a single NVIDIA A100, A40, or V100 GPU.
The angular margin was warmed up at $0.0$ for $10\%$ of the epochs, then increased to $0.2$ as described in \cite{wang2023wespeaker}. Table~\ref{tab:training-config} summarizes the training configuration.
At inference, embeddings were $\ell_2$-normalized before computing cosine similarity. We also applied \ac{AS-norm}~\cite{matejka17_interspeech} with the top-$600$ cohort speakers for score normalization. 


\subsection{Method-specific hyperparameters}
\label{subsec:Sparse-training}
We added our adaptive method to the Bregman framework in \cite{bregman_leon}, namely \texttt{LinBreg} and \texttt{AdaBreg}. We compared our method to the dense and magnitude-pruned baselines. We also compared the performance for the non-adaptive scheme presented in \cite{bregman_leon}, but only for selected experiments.
We excluded normalization layers and bias parameters from the sparse optimization. Meaning, biases and norm layers are never pruned in the pruning case, and trained via \texttt{SGD} when using \texttt{LinBreg} or \texttt{Adam} when using \texttt{AdaBreg}. Table~\ref{tab:bregman-optim} summarizes the settings of the different optimizers in our experiments.

\begin{table}[tb]
    \centering
    \footnotesize
    \caption{
    Optimizer configuration summary. \texttt{lr} indicates the base value and \texttt{wd} is weight decay.
    }
    \label{tab:bregman-optim}
    \renewcommand{\arraystretch}{0.6}
    \setlength{\tabcolsep}{6pt}
    \begin{tabular}{@{} c c l c @{}}
        \toprule
        \textbf{Method} & \textbf{Optimizer} & \hspace{60pt} \textbf{Config} & \textbf{Special Params} \\
        \midrule
        \multirow{2}{*}{Bregman} & \texttt{Linbreg} & \texttt{lr}${=}10^{-1}$, $\lambda_0{=}0.01$ & \multirow{2}{*}{\begin{tabular}{@{}l@{}}ECAPA: conv.: group-norm, linear: $\ell_1$ \\ ResNet34: $\ell_1$ reg. for all layers \end{tabular}} \\
                                 & \texttt{AdaBreg} & \texttt{lr}${=}10^{-2}$, $\lambda_0{=}1.0$ & \\
        \midrule
        \multirow{2}{*}{Dense}  & \texttt{SGD}     & \texttt{lr}${=}10^{-1}$, \texttt{wd}${=}10^{-4}$, Nesterov${=}0.9$ & \multirow{2}{*}{---} \\
        & \texttt{AdamW}   & \texttt{lr}${=}10^{-3}$, \texttt{wd}${=}10^{-4}$ \\
        \midrule
        Pruning                  & \texttt{SGD}     & \texttt{lr}${=}10^{-1}$, \texttt{wd}${=}10^{-4}$, Nesterov${=}0.9$ & Gradual pruning over $10$ epochs \\
        \bottomrule
    \end{tabular}
\end{table}

\textbf{Bregman methods}:
\textbf{(a) Adaptive:} For ECAPA-TDNN, 
we used a group norm \cite{group_sparse} penalty to regularize convolutional layers and $\ell_1$ for linear layers. For ResNet34, we used $\ell_1$ for both convolutional and linear layers, which we found notably better than the group-norm choice used in ECAPA-TDNN. We used learning rates of $0.1$ and $0.01$ for \texttt{LinBreg} and \texttt{AdaBreg}, respectively, and  reduced them by \nicefrac{1}{4} after plateauing with $2$~epochs patience.
Unless mentioned otherwise, we initialized $\lambda_0{=}0.01$ for \texttt{LinBreg} and $1.0$ for \texttt{AdaBreg}, irrespective of the model, dataset, and target sparsity. We used $f{=}50$ and $\alpha{=}1$ for all experiments, except for \texttt{AdaBreg} with ResNet34, where $\alpha$ was set to $0.25$. The damping threshold $\zeta_d{=}0.5\%$ around the target sparsity and $\gamma_{\alpha}$ and $\gamma_f$ are set to $10$ and $2$, respectively. The sparsity acceptance tolerance $\zeta$ was set to $1\%$.
\textbf{(b) Non-adaptive:} 
We only focused on achieving $\mathsf{s}^\ast{=}75\%$. Following empirical tuning, we used the values $\lambda{=}10,\;15$ in \texttt{AdaBreg} and $\lambda{=}0.025,\;0.03$ in \texttt{LinBreg} for ECAPA-TDNN and ResNet34, respectively. The exact sparsity per model and dataset is specified in the results. More information about the selection of $\lambda$ is provided in Appendix~\ref{sec:oracle_finetuning}.

\textbf{Dense baselines}:
We adapted WeSpeaker's \cnceleb recipe to train the dense models. Specifically, we used~\texttt{SGD} with base learning rate of $10^{-1}$, Nesterov momentum${=}0.9$, and weight decay ${=}10^{-4}$.
For comparison with \texttt{AdaBreg}, we trained a dense network using ~\texttt{AdamW} with a base learning rate of $10^{-3}$ and weight decay of $10^{-4}$. We used an exponential learning rate decay with a warm-up epoch.

\textbf{Pruning methods}:
We used magnitude-based pruning as a sparse baseline. We observed that pruning a pretrained model to the exact target sparsity often led to subpar performance. Therefore, we experimented with gradual pruning schedules \cite{gradual_pruning_2018} and observed dramatic improvement, even without pretraining. Hence, we use the gradual schedule and treat the scheduling phase as a pretraining phase. 
For the schedule, $\mathsf{s}_i = 1 - (1 - \mathsf{s}^\ast)^{(\nicefrac{i}{10})}$, where $i$ is the epoch index and $\mathsf{s}_i$ is sparsity at epoch $i$, which spreads pruning over $10$ full epochs.
The \texttt{SGD} optimizer used in dense models was also used to train the pruned ones, but with per-epoch learning-rate decay.

\section{Results}
\label{sec:results}

We refer to our adaptive regularization by the corresponding Bregman optimizer and target sparsity. The non-adaptive baselines are labeled ``fixed`` (denoting a fixed $\lambda$) or their corresponding values.

\subsection{Method validation}

\begin{figure}[tb]
    \centering
    \quad
    \begin{subfigure}[tb]{0.65\textwidth}
    \centering
    \includegraphics[width=\linewidth]{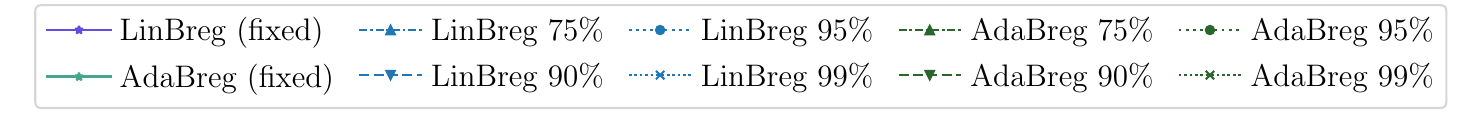}
    \end{subfigure}\\
    \begin{subfigure}[tb]{0.4\textwidth}
    \centering
    \includegraphics[width=\linewidth]{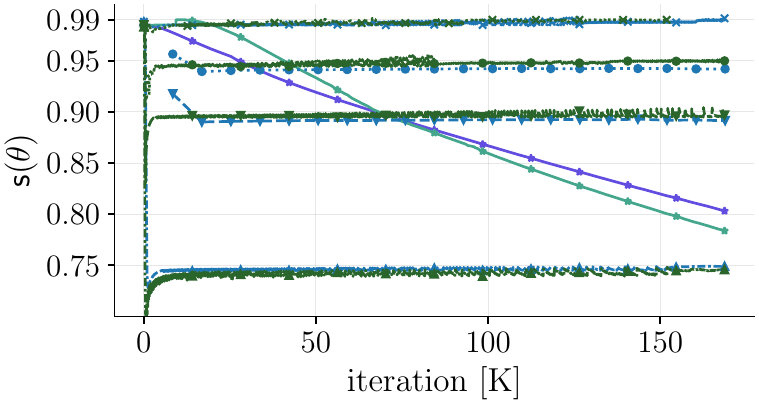}
    \end{subfigure}\qquad
    \begin{subfigure}[tb]{0.4\textwidth}
    \centering
    \includegraphics[width=\linewidth]{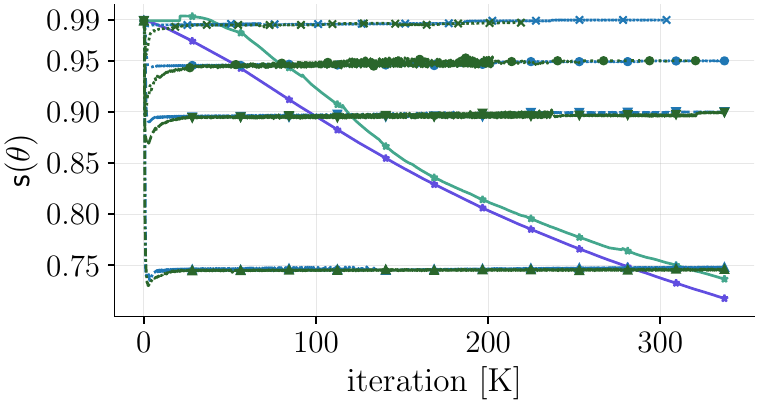}
    \end{subfigure}
    \caption{Sparsity evolution for ECAPA-TDNN (left) and ResNet34 (right) on \voxceleb.}
    \label{fig:internal_curves_both_models}
\end{figure}

\textbf{Reaching $\mathsf{s}^\ast$: }
Figure~\ref{fig:internal_curves_both_models} illustrates the evolution of $\mathsf{s}(\theta)$ for ECAPA-TDNN and ResNet34 on \voxceleb across different $\mathsf{s}^\ast$. Both models and optimizers reach $\mathsf{s}^\ast$ early in training. 
Minor sparsity oscillations occur as a consequence of the $\operatorname{sgn}$ in \eqref{eq:lambda-update}. 
One exception occurred when training ResNet34 with \texttt{AdaBreg} at $\mathsf{s}^\ast{=}95\%$, where oscillations exceeding amplitude $\zeta$ appeared mid-training (Appendix~\ref{sec:exp_results_appendix}, Figure~\ref{fig:internal_curves_cnc}).
Although such an occurrence might be viewed as an instability related to the adaptation, we view it as a signal for a deeper issue related to the incompatibility between \texttt{AdaBreg} and ResNet34. We found that such incompatibility arises from a conflict between the sparsity patterns learned by \texttt{AdaBreg} and the ResNet architecture. Section~\ref{sec:sparsity_patterns} discusses this in more detail.

Another observation from Figure~\ref{fig:internal_curves_both_models} is that \texttt{LinBreg} typically requires $2$ orders of magnitude smaller $\lambda$ to achieve the same sparsity as \texttt{AdaBreg}. This indicates that \texttt{AdaBreg} is biased towards learning weights that have much larger magnitude than what is learned by \texttt{LinBreg} (since it requires larger $\lambda$ to push them to $0$). We believe the bias towards larger weights in  \texttt{AdaBreg} is a result inherited from the Adam-style momentum. This characteristic is unrelated to the adaptation and happens even when using a fixed $\lambda$. In fact, we show in Figure~\ref{fig:l2_norm_coparison} that the Frobenius norm of the network is \emph{higher} in the non-adaptive \texttt{AdaBreg} compared to the adaptive one when measured under the same sparsity.


\begin{figure}[tb]
    \centering
    \quad
    \begin{subfigure}[tb]{0.75\textwidth}
    \centering
    \includegraphics[width=\linewidth]{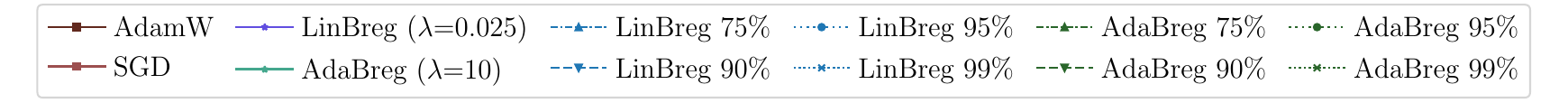}
    \end{subfigure}\\
    \begin{subfigure}[tb]{0.4\textwidth}
    \centering
    \includegraphics[width=\linewidth]{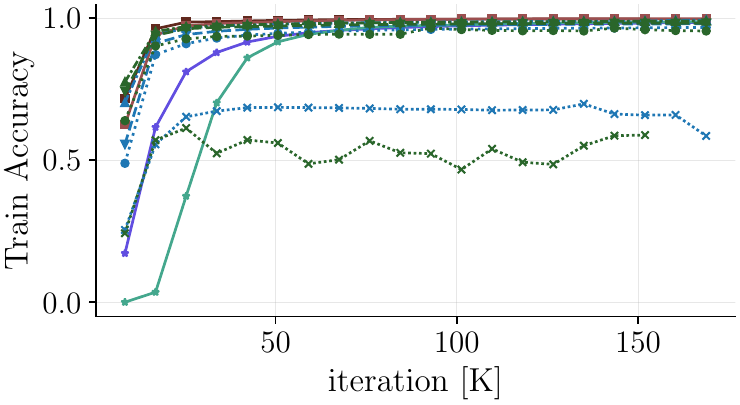}
    \end{subfigure}\qquad
    \begin{subfigure}[tb]{0.4\textwidth}
    \centering
    \includegraphics[width=\linewidth]{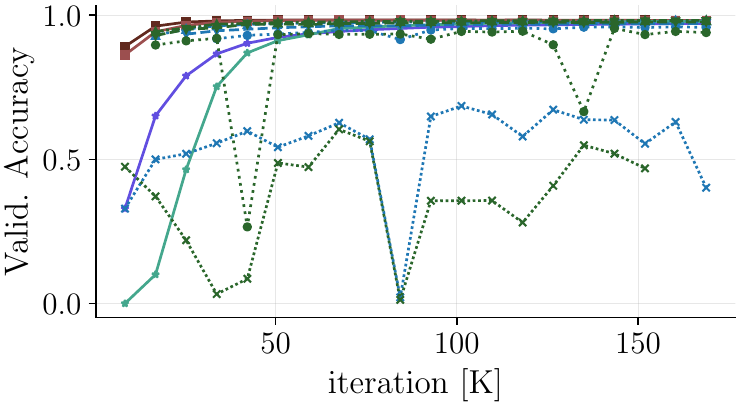}
    \end{subfigure}
    \caption{Convergence of the dense and Bregman-trained ECAPA-TDNN on \voxceleb.}
    \label{fig:convergence_ecapa_voxceleb}
\end{figure}

\textbf{Convergence analysis: }
Here we show that $\lambda$ adaptation does not degrade the convergence. Figure~\ref{fig:convergence_ecapa_voxceleb} shows the convergence of ECAPA-TDNN on \voxceleb compared to the dense baselines. 
The curves indicate that the adaptive models converge \emph{faster} than their non-adaptive counterparts. We attribute the faster convergence to the adaptation mechanism, which quickly reduces sparsity and admits more weights early in training, whereas the non-adaptive approach reduces sparsity much more gradually. Ultimately, both methods achieve comparable final accuracy.

At high sparsity levels, particularly at $99\%$, validation accuracy drops sharply and then quickly recovers, suggesting brittle generalization. Our hypothesis is that the optimizer changes sparse weights as training progresses. At extreme sparsity, this could sever important connections to the validation speakers.
Since the validation set represents only $5\%$ of \voxceleb, these shifts disproportionately degrade validation accuracy without substantially impacting the training loss.

\subsection{Performance on the test sets}
\label{sec:main_results}

\begin{figure}[tb]
    \centering
    \begin{subfigure}[tb]{\textwidth}
        \centering
        \includegraphics[width=0.8\textwidth]{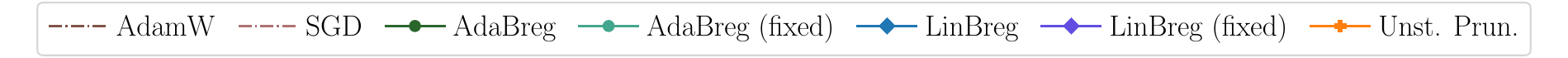}
    \end{subfigure}        
    \begin{subfigure}[tb]{\textwidth}
        \centering
        \includegraphics[width=\textwidth]{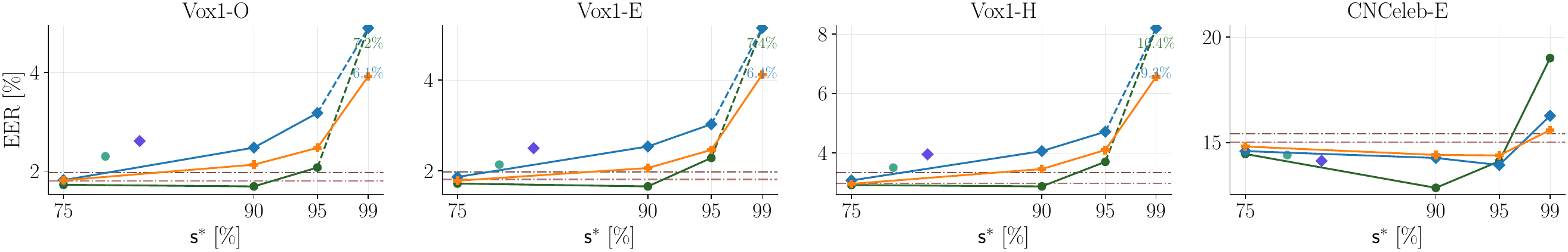}
        \caption{ECAPA-TDNN}
        \label{fig:voxceleb_sparsity_trends_ecapa} 
    \end{subfigure}
    \begin{subfigure}[tb]{\textwidth}
        \centering
        \includegraphics[width=\textwidth]{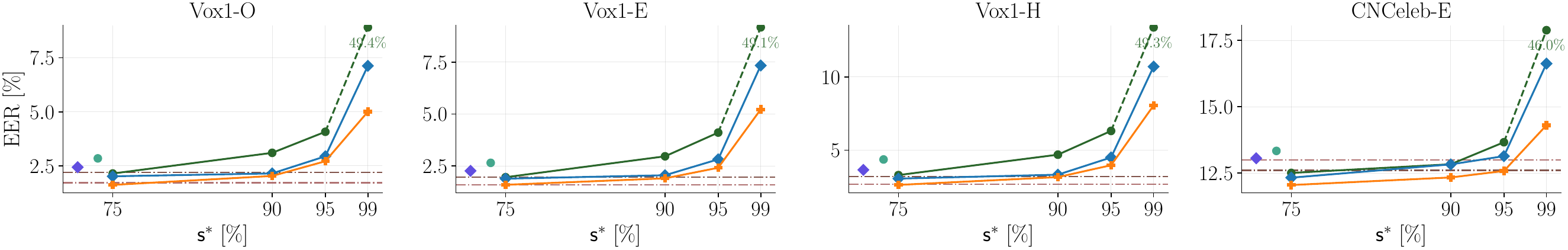}
        \caption{ResNet34}
        \label{fig:voxceleb_sparsity_trends_resnet} 
    \end{subfigure}
    \caption{\Acl{EER} on \voxceleb and \cnceleb-E when training via \voxceleb2 dev. The \cnceleb-E case (rightmost plot) represents the \ac{OOD} scenario. \texttt{AdaBreg} and \texttt{LinBreg} refer to our adaptive method at different target sparsity levels, and ``fixed`` refers to the non-adaptive method.}
    \label{fig:voxceleb_sparsity_trends}
\end{figure}

Figure~\ref{fig:voxceleb_sparsity_trends} shows the \ac{EER} on \voxceleb's test sets and \cnceleb-E, respectively. As the models were trained on \voxceleb, \cnceleb-E represents \ac{OOD} data. Several key trends emerge from the results:

\textbf{Adaptive versus non-adaptive Bregman}: The non-adaptive runs reach sparsity rates of approximately $70$--$80\%$ depending on the model and optimizer. Although models are not directly comparable due to differences in sparsity, non-adaptive baselines are Pareto-dominated by their adaptive counterparts on the in-domain \voxceleb sets. An exception is the \ac{OOD} test with ECAPA-TDNN, where non-adaptive \texttt{LinBreg} is competitive with its adaptive counterpart.
Such an advantage for the adaptive scheme is an important and unexpected finding, as adaptation was designed solely to control sparsity rather than to improve performance. We hypothesize that this is a consequence of prolonged training at extreme sparsity levels during the early iterations, which may lead to worse local minima.

\textbf{Sparse versus dense models}: 
Sparse models are expected to attend to the most robust features in the data and disregard non-robust ones, hence trading off generalization for training accuracy. 
Indeed, our experiments confirm that while dense models achieve the highest training accuracy (Figure~\ref{fig:convergence_ecapa_voxceleb}), sparse models exhibit better generalization on the \ac{OOD} \cnceleb-E set, particularly with ECAPA-TDNN. This reinforces the belief that sparsity encourages the learning of more robust representations.

\textbf{Model-dependent performance}:
Among dense models, ResNet34 slightly outperforms ECAPA-TDNN, especially on \cnceleb-E. Unstructured pruning yields results similar to those of dense models.
Conversely, Bregman optimizers perform better on ECAPA-TDNN than ResNet34, particularly at $99\%$ sparsity.
We attribute this to the interplay between two factors: 1) Compared to ResNet34, ECAPA-TDNN has a significantly lower ratio of classifier weights relative to its backbone architecture ($4.7\%$ vs. $9.8\%$ on \cnceleb and $8.7\%$ vs. $21.4\%$ in \voxceleb); and 2) Bregman optimizers assign greater importance to the classifier, thus consuming more sparsity budget. Consequently, to maintain the target $\mathsf{s}^\ast$, the remaining backbone layers are forced into extreme sparsity (cf.~Section~\ref{sec:sparsity_patterns}).

This reasoning explains why ECAPA-TDNN achieves peak performance among sparse models with \texttt{AdaBreg} at up to $95\%$ sparsity, but degrades sharply at $99\%$ as parameter allocation becomes overly constrained.
For ResNet34, the classifier's larger share forces the remaining layers to adopt even greater sparsity to maintain the target. Hence, the early and intermediate layers become almost completely sparse, resulting in a nearly collapsed model.

The tendency to leave the classifier relatively dense was also observed with unstructured pruning (see Fig.~\ref{fig:sparsity_patterns_fixed}), suggesting vanishing gradients during training. Forcing a sparse classifier by applying single-shot pruning at initialization, rather than gradually over multiple epochs, severely degrades performance. We verified this on ECAPA-TDNN at $\mathsf{s}^\ast{=}90\%$ on \voxceleb, where single-shot pruning yielded a validation accuracy of only $7.5\%$, compared to $96.8\%$ achieved with the gradual schedule.

\textbf{Sparsity role}: ResNet34 shows a clear trade-off between sparsity and performance, but it affects the sparse models differently. In the case of ECAPA-TDNN, the relationship is more nuanced. In general, the same trade-off pattern holds except for \texttt{AdaBreg}, which has the best performance at $90\%$ sparsity. At $99\%$, the increase in sparsity severely degrades \texttt{AdaBreg}. At $75\%$ sparsity, all sparse models are on par with the dense baselines on the in-distribution evaluations.


\begin{figure}[tb]
    \centering
    \begin{subfigure}[tb]{0.75\textwidth}
        \centering
        \includegraphics[width=\textwidth]{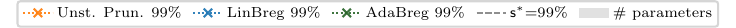}
    \end{subfigure}
    \begin{subfigure}[tb]{0.45\textwidth}
        \centering
        \includegraphics[width=\textwidth]{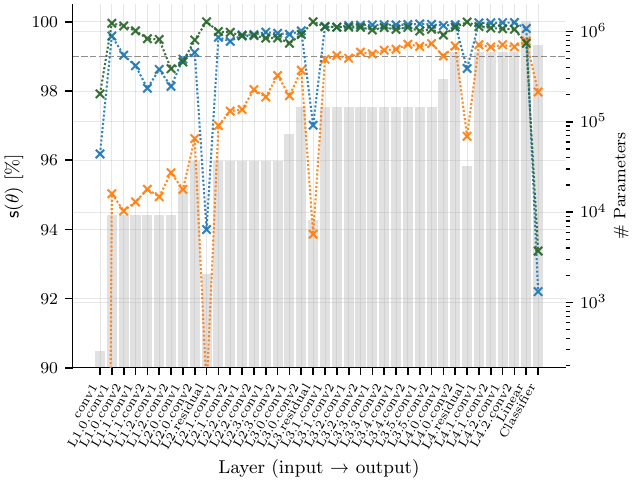}
        \caption{\cnceleb}
    \end{subfigure}
    \begin{subfigure}[tb]{0.45\textwidth}
        \centering
        \includegraphics[width=\textwidth]{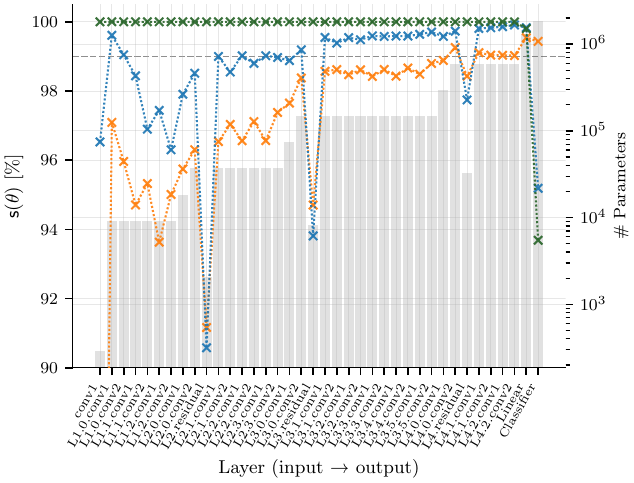}
        \caption{\voxceleb}
    \end{subfigure}
    \caption{Cross-dataset comparison for layer-wise sparsity distribution in ResNet34. Bregman optimizers tend to keep the classifier denser at the expense of other layers, with \texttt{AdaBreg} exhibiting this behavior to a greater extent. For \voxceleb, this results in extreme sparsity for all other layers, since the classifier has more parameters, resulting in a nearly collapsed model.}
    \label{fig:sparsity_patterns}
\end{figure}

\subsection{Learned sparsity patterns}
\label{sec:sparsity_patterns}

Although \texttt{AdaBreg} matches baselines at up to $95\%$ sparsity, it collapses for ResNet34 at $99\%$ on \voxceleb. We show the per-layer allocation in Figure~\ref{fig:sparsity_patterns} for \texttt{AdaBreg}, \texttt{LinBreg}, and gradual pruning.
Both Bregman variants leave the classifier substantially denser than pruning. Under extreme sparsity with a parameter-dense classifier, this bias is doubly damaging: the budget is absorbed by the classifier, forcing intermediate layers to near-complete sparsity and rendering the feature path untrainable, while the classifier itself is discarded at \ac{ASV} inference. The bias is intrinsic to the optimizer and persists even without $\lambda$ adaptation (cf. Fig.~\ref{fig:sparsity_patterns_fixed}).
A simple per-group rescaling $\lambda \leftarrow a\lambda$
($a{=}2$ for the classifier, $a{=}1$ otherwise) recovers the $99\%$ setting (cf.~Table~\ref{tab:before_and_after_2lambda}), with the largest gains for ResNet34 with \texttt{AdaBreg}. Hence, a per-layer sparsity allocation may yield further improvements.

\section{Conclusions}
This work introduces a lightweight adaptive scheme for controlling $\lambda$ in the Bregman sparse optimization framework for both \texttt{AdaBreg} and \texttt{LinBreg} variants. We validated our method on the large-scale \ac{ASV} benchmarks \voxceleb and \cnceleb, and using two well-known models, viz., ECAPA-TDNN and ResNet34.
We showed that our method reliably steers the models towards a user-specified sparsity between $75\%$ and $99\%$. 
In addition to controlling sparsity, our adaptive method achieved faster convergence and on-par or better \ac{EER} compared to the oracle $\lambda$ selection of the non-adaptive scheme.
The adaptation was shown to preserve properties of the original non-adaptive scheme, including robustness to out-of-distribution data and a similar layer-wise sparsity distribution. Our evaluation demonstrated that sparse models trained with Bregman-based optimizers and pruned with a gradual schedule can match the performance of dense models at up to $95\%$ sparsity.

\paragraph{Limitations and future work.} 
The main limitation of our work is the lack of a theoretical analysis of the stability in sparsity oscillations. Deriving bounds on $\alpha$ and $f$ that guarantee decaying sparsity oscillations is an important future direction. Another potential direction is to extend our framework to avoid extreme sparsity in early and intermediate layers when the target sparsity is high.

\paragraph{Broader Impact.} 
By eliminating the per-target $\lambda$ search, our adaptive scheme reduces the training and engineering cost of producing sparse \ac{ASV} models, lowering the energy footprint of model selection and easing deployment on resource-constrained devices. The biometric misuse and demographic fairness considerations inherent to speaker verification are not altered by our optimizer and remain the responsibility of the deployment stage; the sparse models produced here should be evaluated using the same fairness and robustness criteria as their dense counterparts.
\begin{ack}
AA, TR, DT, and EH acknowledge funding by the German Ministry of Research, Technology and Space (BMFTR) under grant agreement No. 16IS24072F (COMFORT). TR acknowledges the support of the Munich Center for Machine Learning.

For the purpose of Open Access, the authors have applied a CC BY public copyright license to any
Author Accepted Manuscript (AAM) version arising from this submission.
\end{ack}

\clearpage
\bibliographystyle{plainnat}
\bibliography{utils/references}

@string{icassp = "Proc. ICASSP"}

@string{interspeech = "Proc. Interspeech"}

@string{ieee-tnnls = "IEEE Trans. Neural Nets. and Lin. Systems"}

@string{icml = "Proc. ICML"}

@string{iclr = "Proc. ICLR"}

@string{iccvw = "ICCV Workshop"}

@string{neurips = "Proc. NeurIPS"}

@string{jmlr = "JMLR"}

@string{apsipa = "Proc. APSIPA"}

@string{ijcnn = "Proc. IJCNN"}

@article{jumper2021alpha_fold,
  title={Highly accurate protein structure prediction with {AlphaFold}},
  author={Jumper, J. and Evans, R. and Pritzel, A. and Green, T. and Figurnov, M. and Ronneberger, O. and Tunyasuvunakool, K. and Bates, R. and {\v{Z}}{\'\i}dek, A. and Potapenko, A. and others},
  journal={Nature},
  volume={596},
  number={7873},
  pages={583--589},
  year={2021},
}

@misc{openai2024gpt4technicalreport,
      title={GPT-4 Technical Report}, 
      author={Josh Achiam and Steven Adler and Sandhini Agarwal and Lama Ahmad and Ilge Akkaya and Florencia Leoni Aleman and Diogo Almeida and Janko Altenschmidt and Sam Altman and Shyamal Anadkat and others},
      year={2024},
      eprint={2303.08774},
      archivePrefix={arXiv},
      primaryClass={cs.CL},
      url={https://arxiv.org/abs/2303.08774}, 
}

@article{dhar2020carbon,
  title={The carbon impact of artificial intelligence},
  author={Dhar, P.},
  journal={Nat. Mach. Intell.},
  volume={2},
  pages={423--425},
  year={2020}
}

@inproceedings{LeCun_optimal_brain_damage,
 author = {LeCun, Yann and Denker, John and Solla, Sara},
 booktitle = neurips,
 title = {Optimal Brain Damage},
 url = {https://proceedings.neurips.cc/paper_files/paper/1989/file/6c9882bbac1c7093bd25041881277658-Paper.pdf},
 volume = {2},
 year = {1989}
}

@article{tibshirani1996regression,
 URL = {http://www.jstor.org/stable/2346178},
 author = {Tibshirani, R.},
 journal = {Jour. Roy. Stat. Soc. Series B},
 number = {1},
 pages = {267--288},
 title = {Regression Shrinkage and Selection via the Lasso},
 volume = {58},
 year = {1996}
}

@inproceedings{han2015learning,
  title={Learning both weights and connections for efficient neural network},
  author={Han, S. and Pool, J. and Tran, J. and Dally, W.},
  booktitle=neurips,
  volume={28},
  year={2015}
}

@article{hoefler2021sparsity,
  title={Sparsity in deep learning: Pruning and growth for efficient inference and training in neural networks},
  author={Hoefler, T. and Alistarh, D. and Ben-Nun, T. and Dryden, N. and Peste, A.},
  journal=jmlr,
  volume={22},
  number={241},
  pages={1--124},
  year={2021}
}

@inproceedings{frankle2018the,
title={The Lottery Ticket Hypothesis: Finding Sparse, Trainable Neural Networks},
author={Jonathan Frankle and Michael Carbin},
booktitle=iclr,
year={2019},
url={https://openreview.net/forum?id=rJl-b3RcF7},
}

@inproceedings{janusz2024oneshot,
title={How Many Does It Take to Prune a Network: Comparing One-Shot vs. Iterative Pruning Regimes},
author={Tomasz Wojnar and Miko{\l}aj Janusz and Luca Benini and Yawei Li and Kamil Adamczewski},
booktitle={Workshop on ML and Compression, Proc. NeurIPS},
year={2024},
}

@inproceedings{huang2016split,
  title={Split LBI: An Iterative Regularization Path with Structural Sparsity},
  author={Huang, Chendi and Sun, Xinwei and Xiong, Jiechao and Yao, Yuan},
  booktitle=neurips,
  pages={3369--3377},
  year={2016}
}

@ARTICLE{Azizan,
  author={Azizan, Navid and Lale, Sahin and Hassibi, Babak},
  journal=ieee-tnnls, 
  title={Stochastic Mirror Descent on Overparameterized Nonlinear Models}, 
  year={2022},
  volume={33},
  number={12},
  pages={7717-7727},
  }

@article{lifted_bregman,
  title={Lifted {Bregman} Training of Neural Networks},
  author={Wang, Xiaoyu and Benning, Martin},
  journal=jmlr,
  volume={24},
  number={232},
  pages={1--51},
  year={2023}
}

@inproceedings{strubell2019energy,
  title={Energy and Policy Considerations for Deep Learning in {NLP}},
  author={Strubell, Emma and Ganesh, Ananya and McCallum, Andrew},
  booktitle={Proc. ACL},
  pages={3645--3650},
  year={2019}
}

@article{luccioni2023estimating,
  title={Estimating the Carbon Footprint of {BLOOM}, a 176B Parameter Language Model},
  author={Luccioni, Alexandra Sasha and Viguier, Sylvain and Ligozat, Anne-Laure},
  journal=jmlr,
  volume={24},
  number={253},
  pages={1--15},
  year={2023}
}

@INPROCEEDINGS{villalobos2022compute,
  author={Sevilla, Jaime and Heim, Lennart and Ho, Anson and Besiroglu, Tamay and Hobbhahn, Marius and Villalobos, Pablo},
  booktitle= ijcnn, 
  title={Compute Trends Across Three Eras of Machine Learning}, 
  year={2022},
  volume={},
  number={},
  pages={1-8},
}

@article{mocanu2018scalable,
  title={Scalable Training of Artificial Neural Networks with Adaptive
         Sparse Connectivity inspired by Network Science},
  author={Mocanu, Decebal Constantin and Mocanu, Elena and Stone, Peter and Nguyen, Phuong H. and Gibescu, Madeleine and Liotta, Antonio},
  journal={Nature Comm.},
  volume={9},
  number={1},
  pages={2383},
  year={2018}
}

@inproceedings{evci2020rigging,
  title={Rigging the Lottery: Making All Tickets Winners},
  author={Evci, Utku and Gale, Trevor and Menick, Jacob and Castro,
          Pablo Samuel and Elsen, Erich},
  booktitle=icml,
  pages={2943--2952},
  year={2020}
}

@InProceedings{ji_advancing_dst,
  title = 	 {Advancing Dynamic Sparse Training by Exploring Optimization Opportunities},
  author =       {Ji, Jie and Li, Gen and Yin, Lu and Qin, Minghai and Yuan, Geng and Guo, Linke and Liu, Shiwei and Ma, Xiaolong},
  booktitle = icml,
  pages = 	 {21606--21619},
  year = 	 {2024},
  volume = 	 {235},
  month = 	 {21--27 Jul},
  url = 	 {https://proceedings.mlr.press/v235/ji24a.html},
}

@misc{dettmers2019sparsenetworksscratchfaster,
      title={Sparse Networks from Scratch: Faster Training without Losing Performance}, 
      author={Tim Dettmers and Luke Zettlemoyer},
      year={2019},
      archivePrefix={arXiv Preprint},
      primaryClass={cs.LG},
      url={https://arxiv.org/abs/1907.04840}, 
}

@book{yudin,
	author = {Nemirovsky, A. S. and Yudin, D. B.},
	publisher = {John Wiley \& Sons},
	title = {Problem Complexity and Method Efficiency in Optimization},
	year = {1983}
    }

@article{kim2016deep,
  title={Deep neural network with weight sparsity control and pre-training  extracts hierarchical features and enhances classification performance: Evidence from whole-brain resting-state functional connectivity patterns of schizophrenia},
  author={Kim, Junghoe and Calhoun, Vince D. and Shim, Eunsoo and Lee, Jong-Hwan},
  journal={NeuroImage},
  volume={124},
  pages={127--146},
  year={2016},
  publisher={Elsevier}
}

@article{shen2024sparse,
  title={Sparse Deep Learning Models with the $\ell_1$ Regularization},
  author={Shen, Lixin and Wang, Rui and Xu, Yuesheng and Yan, Mingsong},
  journal={arXiv preprint arXiv:2408.02801},
  year={2024}
}

@inproceedings{louizos2018learning,
  title={Learning Sparse Neural Networks through $L_0$ Regularization},
  author={Louizos, Christos and Welling, Max and Kingma, Diederik P.},
  booktitle=iclr,
  year={2018}
}

@misc{lunk2026multilevel,
      title={Sparse Training of Neural Networks based on Multilevel Mirror Descent}, 
      author={Yannick Lunk and Sebastian J. Scott and Leon Bungert},
      year={2026},
      eprint={2602.03535},
      archivePrefix={arXiv},
      primaryClass={cs.LG},
      url={https://arxiv.org/abs/2602.03535}, 
}

@inproceedings{HyperSparse2023,
  title={HyperSparse Neural Networks: Shifting Exploration to Exploitation through Adaptive Regularization},
  author={Patrick Glandorf and Timo Kaiser and Bodo Rosenhahn},
  booktitle= iccvw,
  year={2023}
}

@inproceedings{mackay2018selftuning,
title={Self-Tuning Networks: Bilevel Optimization of Hyperparameters using Structured Best-Response Functions},
author={Matthew Mackay and Paul Vicol and Jonathan Lorraine and David Duvenaud and Roger Grosse},
booktitle=iclr,
year={2019},
url={https://openreview.net/forum?id=r1eEG20qKQ},
}

@article{nesterov2009primal,
  title={Primal-dual subgradient methods for convex problems},
  author={Nesterov, Yurii},
  journal={Mathematical programming},
  volume={120},
  number={1},
  pages={221--259},
  year={2009},
  publisher={Springer}
}

@article{yin2008bregman,
  title={{Bregman} iterative algorithms for $\ell_1$-minimization with applications to compressed sensing},
  author={Yin, Wotao and Osher, Stanley and Goldfarb, Donald and Darbon, Jerome},
  journal={SIAM Journal on Imaging sciences},
  volume={1},
  number={1},
  pages={143--168},
  year={2008},
  publisher={SIAM}
}

@article{cai2009linearized,
  title={Linearized {Bregman} iterations for compressed sensing},
  author={Cai, Jian-Feng and Osher, Stanley and Shen, Zuowei},
  journal={Mathematics of computation},
  volume={78},
  number={267},
  pages={1515--1536},
  year={2009}
}

@article{benning2017choose,
  title={Choose your path wisely: Gradient descent in a {Bregman} distance framework},
  author={Benning, Martin and Betcke, Marta M and Ehrhardt, Matthias J and Sch{\"o}nlieb, Carola-Bibiane},
  journal={arXiv preprint arXiv:1712.04045},
  year={2017}
}

@book{rockafellar1998variational,
  title={Variational Analysis},
  author={Rockafellar, R Tyrrell and Wets, Roger JB},
  year={1998},
  publisher={Springer}
}

@article{martin2025variable,
  title={Variable {Bregman} majorization-minimization algorithm and its application to Dirichlet maximum likelihood estimation},
  author={Martin, S{\'e}gol{\`e}ne and Pesquet, Jean-Christophe and Steidl, Gabriele and Ayed, Ismail Ben},
  journal={arXiv preprint arXiv:2501.07306},
  year={2025}
}

@article{zhang2010bregmanized,
  title={Bregmanized nonlocal regularization for deconvolution and sparse reconstruction},
  author={Zhang, Xiaoqun and Burger, Martin and Bresson, Xavier and Osher, Stanley},
  journal={SIAM journal on imaging sciences},
  volume={3},
  number={3},
  pages={253--276},
  year={2010},
  publisher={SIAM}
}

@article{benfenati2013inexact,
  title={Inexact {Bregman} iteration with an application to Poisson data reconstruction},
  author={Benfenati, Alessandro and Ruggiero, Valeria},
  journal={Inverse Problems},
  volume={29},
  number={6},
  pages={065016},
  year={2013},
  publisher={IOP Publishing}
}

@inproceedings{kingma2015adam,
  title={Adam: A Method for Stochastic Optimization},
  author={Kingma, Diederik P. and Ba, Jimmy},
  booktitle=iclr,
  year={2015},
}

@article{adly2026variable,
  title={Variable {Bregman} Majorization-Minimization algorithms for nonconvex nonsmooth optimization, with application to Poisson imaging},
  author={Adly, Maxence and Chazottes, Alix and Chouzenoux, {\'E}milie and Pesquet, Jean-Christophe and Sureau, Florent},
  journal={arXiv preprint arXiv:2604.12829},
  year={2026}
}

@book{Bauschke2017,
  title = {Convex Analysis and Monotone Operator Theory in Hilbert Spaces},
  ISBN = {9783319483115},
  ISSN = {2197-4152},
  url = {http://dx.doi.org/10.1007/978-3-319-48311-5},
  DOI = {10.1007/978-3-319-48311-5},
  journal = {CMS Books in Mathematics},
  publisher = {Springer International Publishing},
  author = {Bauschke,  Heinz H. and Combettes,  Patrick L.},
  year = {2017}
}

@article{bregman_leon,
  author  = {Leon Bungert and Tim Roith and Daniel Tenbrinck and Martin Burger},
  title   = {A {Bregman} Learning Framework for Sparse Neural Networks},
  journal = {Journal of Machine Learning Research},
  year    = {2022},
  volume  = {23},
  number  = {192},
  pages   = {1--43},
  url     = {http://jmlr.org/papers/v23/21-0545.html}
}

@article{group_sparse,
title = {Group sparse regularization for deep neural networks},
journal = {Neurocomputing},
volume = {241},
pages = {81-89},
year = {2017},
issn = {0925-2312},
url = {https://www.sciencedirect.com/science/article/pii/S0925231217302990},
author = {Simone Scardapane and Danilo Comminiello and Amir Hussain and Aurelio Uncini},
}

@article{osher2005iterative,
  title={An iterative regularization method for total variation-based image restoration},
  author={Osher, Stanley and Burger, Martin and Goldfarb, Donald and Xu, Jinjun and Yin, Wotao},
  journal={Multiscale Modeling \& Simulation},
  volume={4},
  number={2},
  pages={460--489},
  year={2005},
  publisher={SIAM}
}

@inproceedings{gunasekar2018characterizing,
  title={Characterizing implicit bias in terms of optimization geometry},
  author={Gunasekar, Suriya and Lee, Jason and Soudry, Daniel and Srebro, Nathan},
  booktitle={International Conference on Machine Learning},
  pages={1832--1841},
  year={2018},
  organization={PMLR}
}

@article{candes2006stable,
  title={Stable signal recovery from incomplete and inaccurate measurements},
  author={Candes, Emmanuel J and Romberg, Justin K and Tao, Terence},
  journal={Communications on Pure and Applied Mathematics: A Journal Issued by the Courant Institute of Mathematical Sciences},
  volume={59},
  number={8},
  pages={1207--1223},
  year={2006},
  publisher={Wiley Online Library}
}

@misc{gradual_pruning_2018,
title={To Prune, or Not to Prune: Exploring the Efficacy of Pruning for Model Compression},
author={Michael H. Zhu and Suyog Gupta},
year={2018},
url={https://openreview.net/forum?id=S1lN69AT-},
}

@INPROCEEDINGS{voxceleb,
  author       = {Chung, J.~S. and Nagrani, A. and Zisserman, A.},
  title        = {{VoxCeleb2: Deep Speaker Recognition}},
  booktitle    = interspeech,
  year         = "2018",
  pages={1086-1090}
}

@article{cnceleb,
    title = {CN-Celeb: Multi-genre speaker recognition},
    journal = {Speech Communication},
    volume = {137},
    pages = {77-91},
    year = {2022},
    issn = {0167-6393},
    author = {Lantian Li and Ruiqi Liu and Jiawen Kang and Yue Fan and Hao Cui and Yunqi Cai and Ravichander Vipperla and Thomas Fang Zheng and Dong Wang},
}

@inproceedings{desplanques_ecapa-tdnn_2020,
	title = {{ECAPA}-{TDNN}: {Emphasized} {Channel} {Attention}, {Propagation} and {Aggregation} in {TDNN} {Based} {Speaker} {Verification}},
	booktitle = interspeech,
	author = {Desplanques, Brecht and Thienpondt, Jenthe and Demuynck, Kris},
	year = {2020},
	pages = {3830--3834},
}

@inproceedings{wang2023wespeaker,
  title={Wespeaker: A research and production oriented speaker embedding learning toolkit},
  author={Wang, Hongji and Liang, Chengdong and Wang, Shuai and Chen, Zhengyang and Zhang, Binbin and Xiang, Xu and Deng, Yanlei and Qian, Yanmin},
  booktitle=icassp,
  pages={1--5},
  year={2023},
  organization={IEEE}
}

@inproceedings{matejka17_interspeech,
  title     = {{Analysis of Score Normalization in Multilingual Speaker Recognition}},
  author    = {Pavel Matějka and Ondřej Novotný and Oldřich Plchot and Lukáš Burget and Mireia Diez Sánchez and Jan Černocký},
  year      = {2017},
  booktitle = interspeech,
  pages     = {1567--1571},
}

@INPROCEEDINGS{aam_softmax,
  author={Xiang, Xu and Wang, Shuai and Huang, Houjun and Qian, Yanmin and Yu, Kai},
  booktitle=apsipa, 
  title={Margin Matters: Towards More Discriminative Deep Neural Network Embeddings for Speaker Recognition}, 
  year={2019},
  pages={1652-1656},
  }

\clearpage
\appendix


\section{Oracle Finetuning}
\label{sec:oracle_finetuning}
Manually identifying $\lambda$ that leads to $\mathsf{s}^\ast$ is a tedious process, especially at scale. We empirically observed that finding such a $\lambda$ depends on numerous factors, including the model architecture, the number of iterations (i.e., the dataset size), and hyperparameter choices such as the learning rate.  

Beyond the tuning cost, a practical issue arises from the observation that network sparsity consistently decreases during training. Because of this dynamic, it is unlikely that a $\lambda$ can satisfy the target sparsity $\mathsf{s}^\ast$ throughout the entire training process. Instead, choosing a certain value of $\lambda$ shifts the training window where the model intersects with the sparsity constraint in \eqref{eq:sparsity_tolerance}. Consequently, multiple valid $\lambda$ values may exist, each achieving $\mathsf{s}^\ast$ at different training stages (e.g., during epochs 2–5 rather than epochs 19–22).

Therefore, in practice, the task boils down to determining the value of $\lambda$ that satisfies the constraint \eqref{eq:sparsity_tolerance} at a certain training iteration. A sensible approach is to calibrate this timing so that the constraint is met neither too early, before meaningful representations are learned, nor too late, when the risk of overfitting increases. We applied this exact rationale when establishing our fixed $\lambda$ values.


\section{Properties of the Proposed Algorithm}
\label{sec:proposed_properties}

We highlight the following properties of the proposed algorithm:
\begin{enumerate}[label=(\Roman*), leftmargin=*, wide]
  \item \textbf{Asymptotic consistency:} Between consecutive $\lambda$-updates the Bregman iterations process $B$ training examples, where $B \in \mathbb{N}$ is the mini-batch size. This provides enough gradient signal for $\theta^{(k)}$ to make meaningful progress towards a local minimizer of $\mathcal{L}^{(k)}$ before $\lambda$ changes. As $f\to\infty$ the formulation approaches the non-adaptive scheme defined by the Bregman iterations.
  
  \item \textbf{Convergence to Bregman iterations:} Assuming sparsity converges to $\mathsf{s}^\ast$ during training, $\epsilon^{(k)}{=0}$ represents an equilibrium state for which $\lambda^{(k+1)}{=}\lambda^{(k)}$. This renders $\phi_{\lambda^{(k)}}{=}\phi_{\lambda^{(k+f)}}{=}\phi$, which reduces the solution of the adaptive scheme to the classical Bregman iterations.  
  
\item \textbf{Bounded absolute relative change:}
  From the update rule~\eqref{eq:lambda-update}, the relative change is
  \begin{equation}
    \frac{|\lambda^{k+1}-\lambda^{(k)}|}{\lambda^{(k)}} = \bigl|(1+\alpha|\epsilon^{(k)}|)^{\operatorname{sgn}(\epsilon^{(k)})}- 1\bigr|.
  \end{equation}

  If we consider the $\operatorname{sgn}$ to be either $-1$ or $+1$, the following is obtained 
  \begin{empheq}[left={\displaystyle \frac{|\lambda^{(k+1)}-\lambda^{(k)}|}{\lambda^{(k)}} = \empheqlbrace}]{align}
      &(1+\alpha|\epsilon^{(k)}|) - 1 \;=\; \alpha|\epsilon^{(k)}| = \alpha\epsilon^{(k)}, && \text{if } \mathsf{s}^{\ast} \geq \epsilon^{(k)} \geq 0 \label{eq:lambda_bounds_1} \\[1ex]
    &\bigl|(1+\alpha|\epsilon^{(k)}|)^{-1} - 1\bigr| \;=\; \frac{\displaystyle \alpha|\epsilon^{(k)}|}{\displaystyle 1+\alpha|\epsilon^{(k)}|}, && \text{if } 0 > \epsilon^{(k)} \geq \mathsf{s}^{\ast}-1. \label{eq:lambda_bounds_2}
  \end{empheq}
  By choosing $\displaystyle 0 < \alpha$, \eqref{eq:lambda_bounds_1} always upper bounds \eqref{eq:lambda_bounds_2}. 
  Thus, $\alpha\epsilon^{(k)}$ is an upper bound that is tight in \eqref{eq:lambda_bounds_1}, i.e., low-sparsity regime. Note that the two factors in $\alpha\epsilon^{(k)}$ are first order and can be made arbitrarily small by driving $\alpha$ to $0$. 
  We note that~\eqref{eq:lambda_bounds_1} also bounds the shift in the proximal threshold, which we define as |$\lambda^{(k+1)}-\lambda^{(k)}|$, by $\alpha\epsilon^{(k)}\lambda^{(k)}$. This shift bounds how far the proximal threshold moves between updates; the number of parameters crossing the threshold additionally depends on the local density of $\bigl|p_i^{(k)}\bigr|$ near $\lambda^{(k)}$.
  
  The high-sparsity initialization of $\theta^{(0)}$, typically at $99\%$ \cite{bregman_leon}, likely results in $\epsilon^{(0)} < 0$, effectively biasing the training towards the regime in \eqref{eq:lambda_bounds_2}. The asymmetric nature of the updates decreases $\lambda^{(k)}$ in small, controlled steps from $\mathsf{s}(\theta^{(0)})$ down to $\mathsf{s}^\ast$ at the start of training, and only takes larger steps to increase $\lambda$ when the model drifts back toward dense, preventing the sparsity rate from collapsing. 
  Furthermore, the asymmetry has minimal impact on the oscillation of $\epsilon^{(k)}$ around $0$ since $\displaystyle \nicefrac{\alpha|\epsilon^{(k)}|}{1+\alpha|\epsilon^{(k)}|} \approx \alpha\epsilon^{(k)}$ for $\epsilon^{(k)} \approx 0$.
\end{enumerate}


\section{Subgradient calculus}

\subsection{Proof of Lemma~\ref{lem:lossdecay}}\label{app:proof}

The $L$-smoothness of $\loss$ implies
\begin{align*}
\loss(\theta^{(k+1)}) - &\loss(\theta^{(k)}) - \frac{L}{2}| \theta^{(k+1)} - \theta^{(k)} |^2\leq 
\langle \nabla \loss(\theta^{(k)}), \theta^{(k+1)} - \theta^{(k)} \rangle
\\&= 
-\frac{1}{\tau}
\langle  p^{(k+1)} - p^{(k)}, \theta^{(k+1)} - \theta^{(k)} \rangle.
\end{align*}
Now we have that
\begin{align*}
p^{(k)} \in \partial \EN_{\lambda^{(k-1)}}(\theta^{(k)}) = 
\theta^{(k)} + \lambda^{(k-1)} g^{(k)}
\end{align*}
where $g^{(k)} \in \partial|\dummy|_1(\theta^{(k)})$ and thus $\langle g^{(k)}, \theta^{(k)}\rangle = |\theta^{(k)}|_1$, which yields
\begin{align*}
\langle  p^{(k+1)} - p^{(k)}, \theta^{(k+1)} - \theta^{(k)} \rangle = 
|\theta^{(k+1)} - \theta^{(k)}|^2 + \langle \lambda^{(k)} g^{(k+1)} - \lambda^{(k-1)} g^{(k)}, \theta^{(k+1)} - \theta^{(k)}\rangle.
\end{align*}
and further
\begin{align*}
&\langle \lambda^{(k)} g^{(k+1)} - \lambda^{(k-1)} g^{(k)}, \theta^{(k+1)} - \theta^{(k)}\rangle\\
&=
\lambda^{(k)} |\theta^{(k+1)}|_1 + \lambda^{(k-1)}  |\theta^{(k)}|_1 - 
\lambda^{(k)}\langle  g^{(k+1)}, \theta^{(k)}\rangle-
\lambda^{(k-1)}\langle  g^{(k)}, \theta^{(k+1)}\rangle
\\&\geq
\lambda^{(k)} |\theta^{(k+1)}|_1 + \lambda^{(k-1)}  |\theta^{(k)}|_1 - 
\lambda^{(k)} |\theta^{(k)}|_1-
\lambda^{(k-1)}|\theta^{(k+1)}|_1
\\
&=
(\lambda^{(k)} - \lambda^{(k-1)})
(|\theta^{(k+1)}|_1 - |\theta^{(k)}|_1).
\end{align*}
Finally, we obtain
\begin{align*}
\loss(\theta^{(k+1)}) - \loss(\theta^{(k)}) &- \frac{L}{2}| \theta^{(k+1)} - \theta^{(k)} |^2\leq \\
&-\frac{1}{\tau} |\theta^{(k+1)} - \theta^{(k)}|^2 -
\frac{\lambda^{(k)} - \lambda^{(k-1)}}{\tau}(|\theta^{(k+1)}|_1 - |\theta^{(k)}|_1),
\end{align*}
which gives the desired statement.

\subsection{Subgradient correction}\label{app:corr}

We briefly comment on the possibility of correcting the $p^{(k+1)}$ in \eqref{eq:mdup}, such that it is a subgradient at the next iteration again. We consider the elastic net functional
\begin{align*}
\operatorname{EN}_{\lambda} = 
\frac{1}{2}|\dummy|_2^2 + \lambda |\dummy|_1.
\end{align*}
Now, if $p\in \partial\operatorname{EN}_\lambda(\theta)$ for some $\lambda >0$, for a different $\tilde{\lambda}>0$ we have 
\begin{align*}
p \in \theta + \lambda\partial |\dummy|_1(\theta) \quad&\Rightarrow\quad \frac{p-\theta}{\lambda} \in \partial|\dummy|_1(\theta) 
\quad\Rightarrow\quad \frac{\tilde{\lambda}}{\lambda} (p-\theta) + \theta \in \theta +  \tilde{\lambda}\partial|\dummy|_1(\theta) \\
&\Rightarrow
\frac{\tilde{\lambda}}{\lambda}\, p + 
\left(1 - \frac{\tilde{\lambda}}{\lambda}\right) \theta \in 
\partial\operatorname{EN}_{\tilde{\lambda}}(\theta).
\end{align*}
Beyond that, we note that at indices $i$ where $\theta_i=0$ we can perform a simpler correction, by simply clipping to the valid range $[-\tilde{\lambda}, \tilde{\lambda}]$ which yields the subgradient correction $C(p, \theta)\in\R^d$ with
\begin{align*}
C(p, \theta, \lambda,\tilde{\lambda})_i :=
\begin{cases}
\frac{\tilde{\lambda}}{\lambda}\, p_i + 
\left(1 - \frac{\tilde{\lambda}}{\lambda}\right) \theta_i &\text{ if } |\theta_i| > 0,\\
\min\{\max\{p_i, -\tilde{\lambda}\}, \tilde{\lambda}\} &\text{ else}.
\end{cases}
\end{align*}
for $i=1,\ldots,d$. This suggests modifying the update in \eqref{eq:md} as follows
\begin{align*}
\tilde{p}^{(k+1)} &:=p^{(k)} - \tau \nabla\loss(\theta^{(k)})\\
\theta^{(k+1)} &:= 
\nabla(\phi^{(k)})^*
\left(\tilde{p}^{(k+1)} \right) \qquad&&\Rightarrow\qquad 
\tilde{p}^{(k+1)}\in\partial\EN_{\lambda^{(k)}}(\theta^{(k+1)}) \\
p^{(k+1)} &:= C(p^{(k)}, \theta^{(k+1)},\lambda^{(k)},\lambda^{(k+1)})
\qquad&&\Rightarrow\qquad 
p^{(k+1)}\in\partial\EN_{\lambda^{(k+1)}}(\theta^{(k+1)}).
\end{align*}
However, we observed empirically in Appendix~\ref{sec:subgradcorr_proxrescale}, that this scheme does not improve performance of the algorithm and in fact can sometimes hurt the desired sparsity constraints and thus we did not pursue this direction in the main results. An intuitive explanation for this would be that when, for example, $|\theta^{(k)}|_0 > \mathsf{a}^*$, we want to increase $\lambda^{(k+1)}$ such that more components of the current subgradient $p^{(k+1)}$ are set to zero by the soft-thresholding operation. However, the rescaling by $\frac{\lambda^{(k+1)}}{\lambda^{(k)}}$ increases the magnitude of the subgradients such that they can again be above the threshold $\lambda^{(k+1)}$. This means in this case the correction can partially negate the desired effect of the parameter update.

\subsection{Prox rescaling}\label{app:proxrescale}

In this section, we compare the parameter adaptation strategy in the original averaging algorithm of~\cite{nesterov2009primal} to our scheme in \eqref{eq:mdup}. While we only update the weight of the $\ell_1$-term in the elastic net, \cite{nesterov2009primal} considers an adaptive parameter $\beta^{(k)}$, which scales the whole functional $\EN_\lambda$. Here, we note that for $\lambda=1$ (see e.g. \cite[Prop. 11.3]{rockafellar1998variational}),
\begin{align*}
\nabla (\beta\EN_1)^*(z) &= \argmax_{\theta} \langle \theta, z \rangle - \beta \EN(\theta) = \argmin_{\theta} -\langle \theta, z/\beta \rangle + \frac{1}{2}|\theta|_2^2 + |\theta|_1\\
&=
\argmin_{\theta} \frac{1}{2}|\theta - z/{\beta}|^2_2 +  |\theta|_1 = 
\prox_{|\dummy|_1}(z/\beta).
\end{align*}
Furthermore, since $|\dummy|_1$ is positively $1$-homogeneous, we obtain
\begin{align*}
\prox_{|\dummy|_1}(z/\beta)&=
\argmin_{\theta} \frac{1}{2}|\theta - z/{\beta}|^2_2 +  |\theta|_1 = 
\argmin_{\theta} \frac{1}{2}|\beta \theta - z|^2_2 +  \beta |\beta\theta|_1\\
&= 
\frac{1}{\beta}\argmin_{\rho} \frac{1}{2}|\rho - z|^2_2 +  \beta |\rho|_1 
=
\frac{1}{\beta}\prox_{\beta|\dummy|_1}(z).
\end{align*}
This leads to a prox-rescaling in \eqref{eq:mdup}, which yields the update
\begin{align*}
p^{(k+1)} &:=p^{(k)} - \tau \nabla\loss(\theta^{(k)})\\
\theta^{(k+1)} &:= \frac{1}{\beta^{(k)}}\prox_{\beta^{(k)}|\dummy|_1}(p^{(k+1)}) 
%
\end{align*}
As with the subgradient correction, we explore this option empirically in Appendix~\ref{sec:subgradcorr_proxrescale} and observe slightly worse performance.


\section{Experimental Results}
\label{sec:exp_results_appendix}

\subsection{Method validation}

\begin{figure}[tb]
    \centering
    \qquad
    \begin{subfigure}[tb]{0.75\textwidth}
    \centering
    \includegraphics[width=\linewidth]{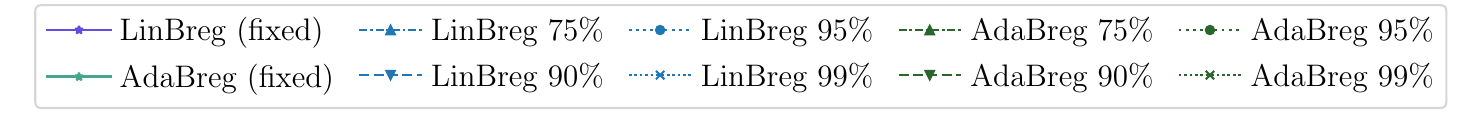}
    \end{subfigure}\\
    \begin{subfigure}[tb]{0.4\textwidth}
    \centering
    \includegraphics[width=\linewidth]{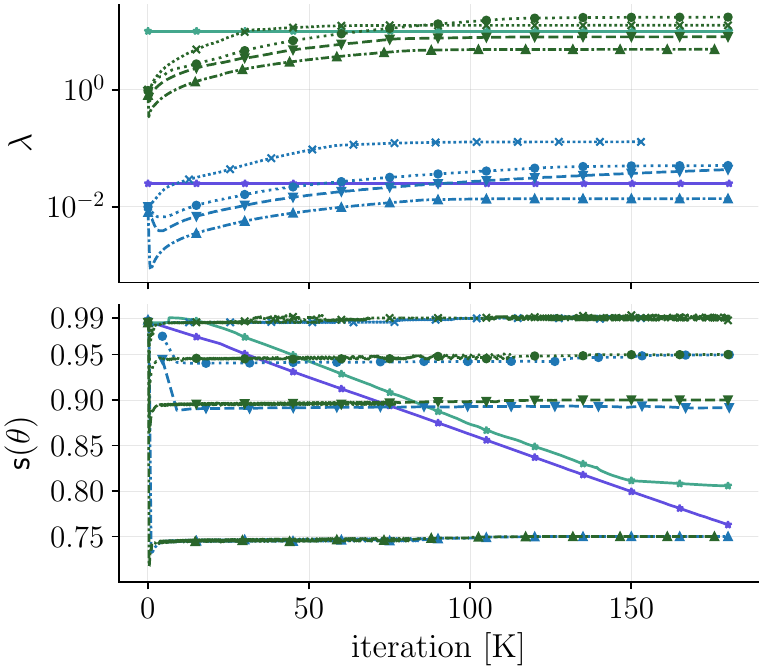}
    \caption{ECAPA-TDNN}
    \end{subfigure}\qquad
    \begin{subfigure}[tb]{0.4\textwidth}
    \centering
    \includegraphics[width=\linewidth]{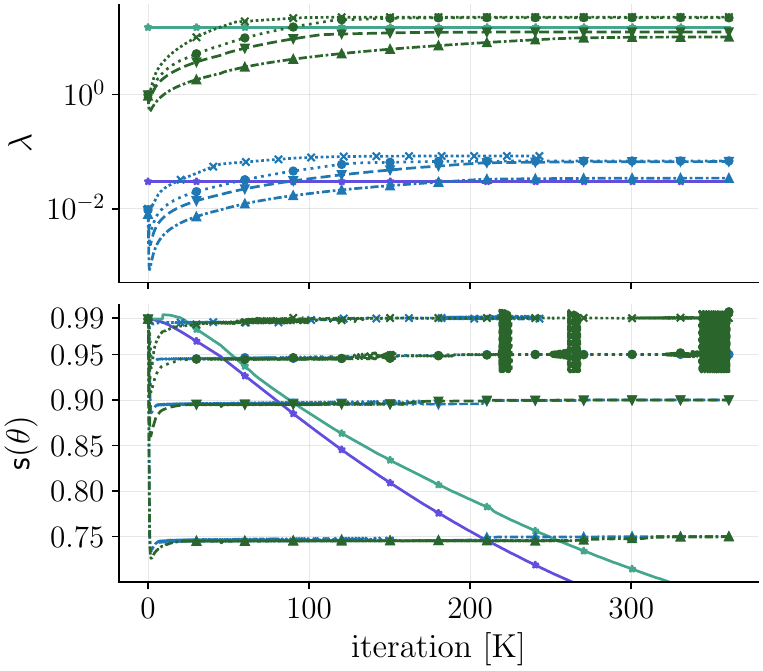}
    \caption{ResNet34}
    \end{subfigure}
    \caption{$\lambda$ and sparsity profiles on \cnceleb. ResNet34 sustains large-amplitude oscillations for $\mathsf{s}^\ast{=}95\%$ at an advanced training stage.}
    \label{fig:internal_curves_cnc}
\end{figure}

Here we show the evolution of sparsity and $\lambda$ throughout training on \cnceleb. The ResNet34 case in Figure \ref{fig:internal_curves_cnc} presents an intriguing anomaly; $\mathsf{s}(\theta)$ oscillations with amplitude larger than $\zeta$, indicating that the scheduler struggles to find $\lambda$ that satisfies $\mathsf{s}^\ast$. We believe this is explained by the bias in sparsity allocation exhibited by \texttt{AdaBreg} as discussed in Section \ref{sec:sparsity_patterns}. We still do not have a sufficient explanation for why this happens in the particular case of \cnceleb at $95\%$ sparsity, or in the other cases.

\subsection{Evaluation on \cnceleb}
\begin{figure}[tb]
    \centering
    \includegraphics[width=0.75\textwidth]{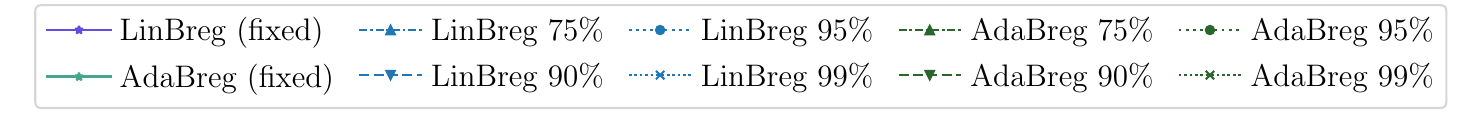}
    \begin{subfigure}[b]{\textwidth}
        \centering
        \includegraphics[width=0.45\textwidth]{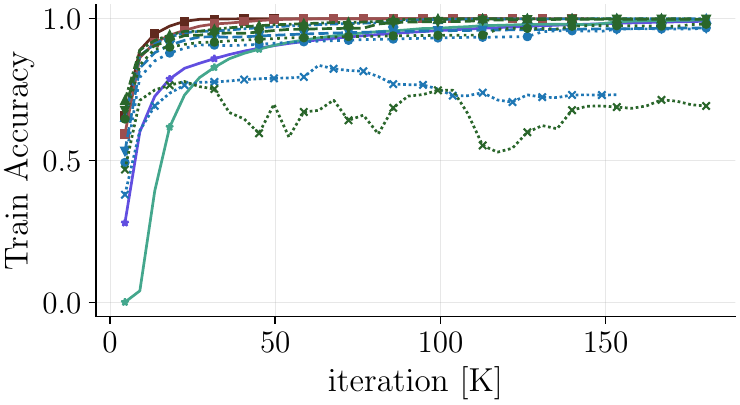}\qquad
        \includegraphics[width=0.45\textwidth]{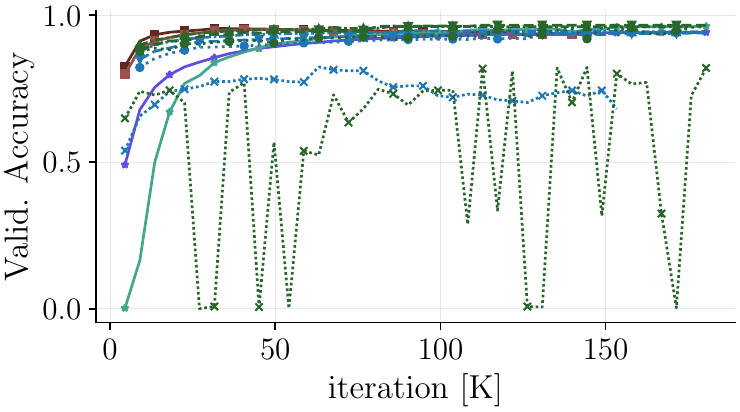}
        \caption{ECAPA-TDNN}
    \end{subfigure}
    \begin{subfigure}[b]{\textwidth}
        \centering
        \includegraphics[width=0.45\textwidth]{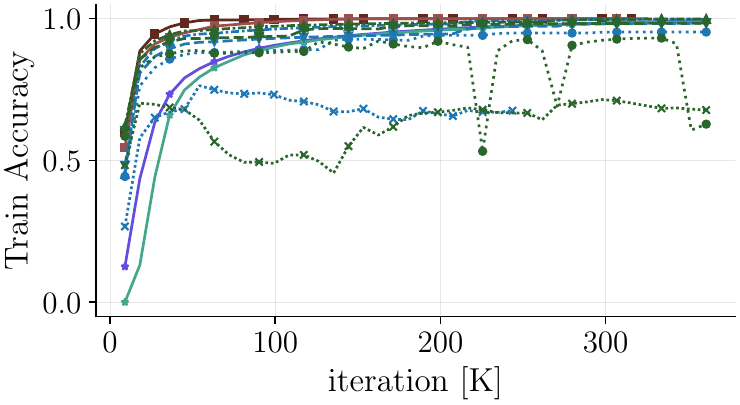}\qquad
        \includegraphics[width=0.45\textwidth]{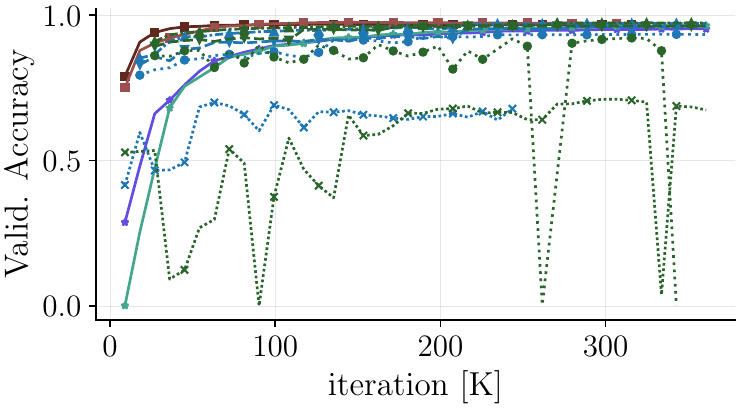}
        \caption{ResNet34}
    \end{subfigure}
    \caption{Convergence of ECAPA-TDNN and ResNet34 on \cnceleb-D.}
    \label{fig:convergence_cnceleb}
\end{figure}

\begin{figure}[tb]
    \centering    
    \begin{subfigure}[tb]{0.49\textwidth}
        \centering
        \includegraphics[width=\textwidth]{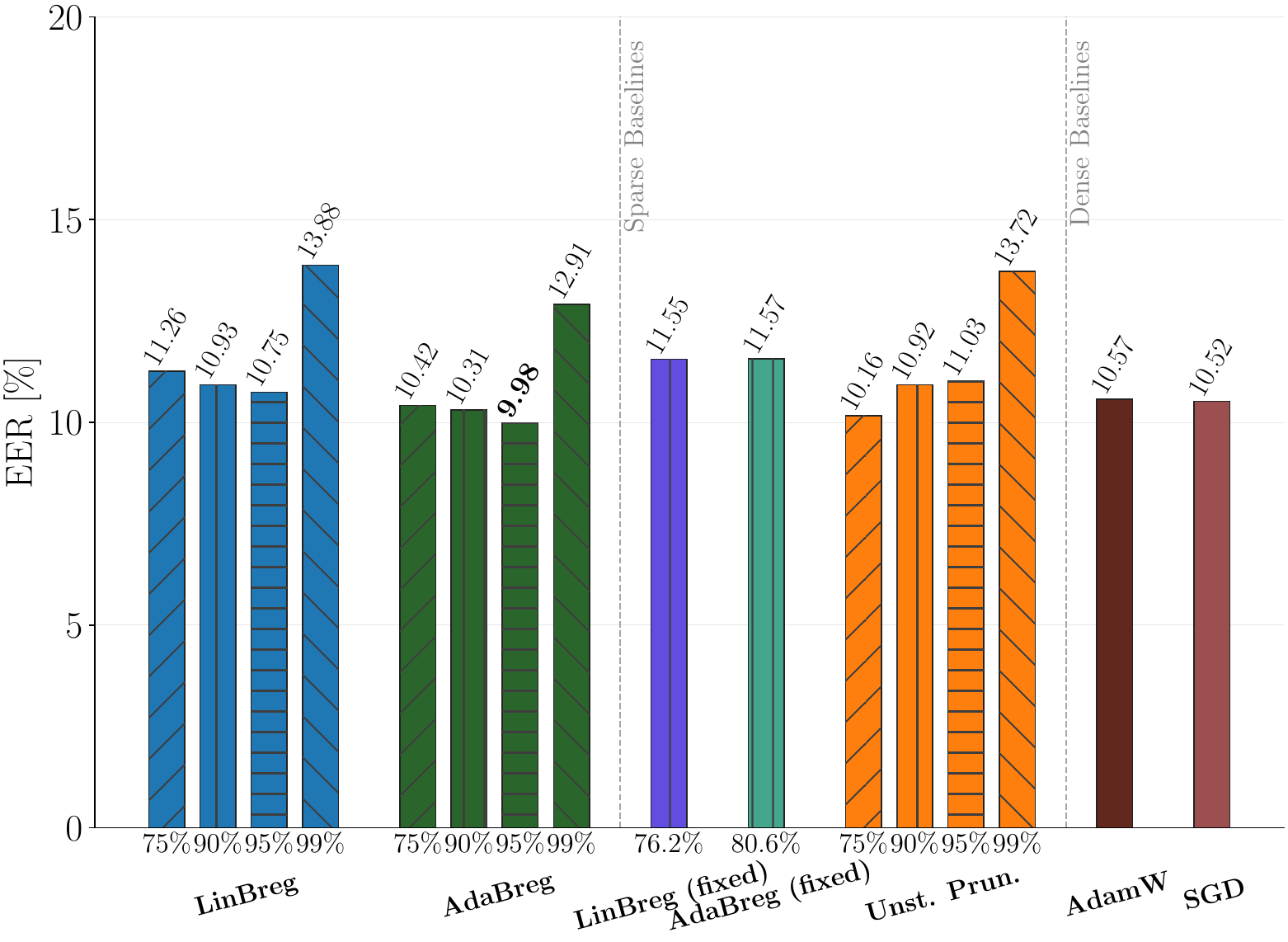}
        \caption{ECAPA-TDNN}
        \label{fig:cnceleb_sparsity_trends_ecapa} 
    \end{subfigure}
    \begin{subfigure}[tb]{0.49\textwidth}
        \centering
        \includegraphics[width=\linewidth]{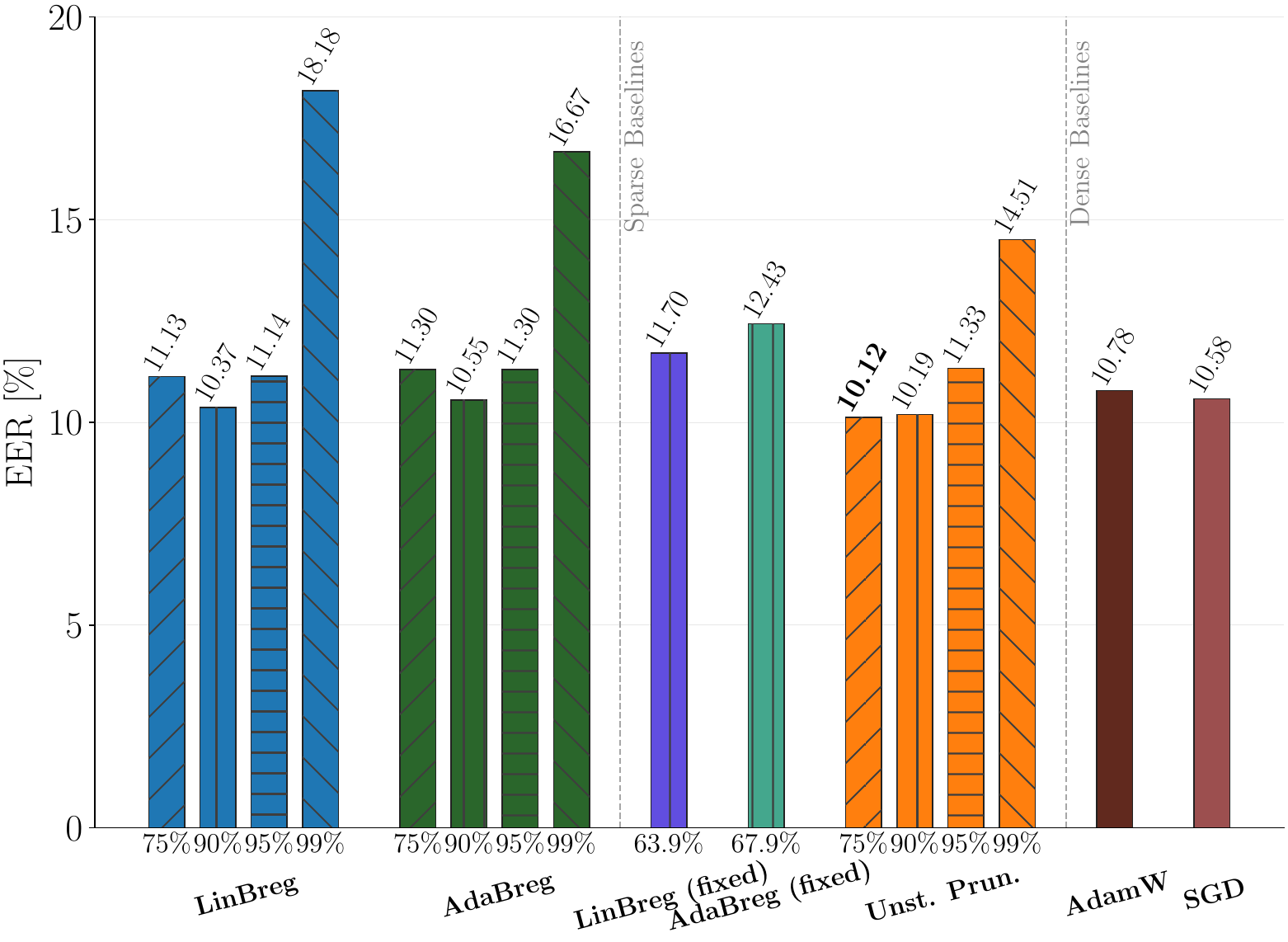}
    \caption{ResNet34}
    \label{fig:cnceleb_sparsity_trends_resnet} 
    \end{subfigure}
    \caption{\ac{EER} on \cnceleb-E when training on \cnceleb-D.}
    \label{fig:cnceleb_sparsity_trends}
\end{figure}

\begin{figure}[tb]
    \centering
    \includegraphics[width=0.85\linewidth]{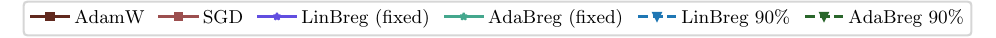}\\
    \begin{subfigure}[tb]{0.45\linewidth}
    \centering
        \includegraphics[width=\linewidth]{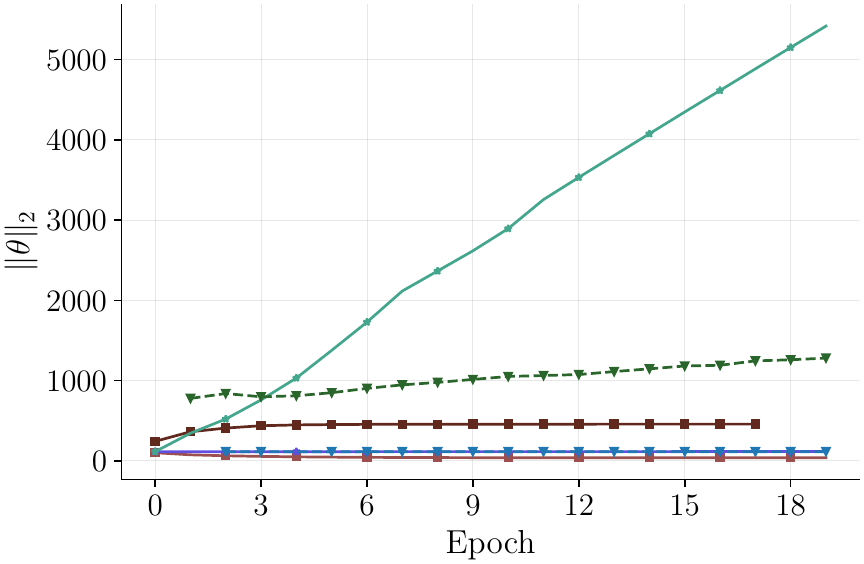}
        \caption{ECAPA-TDNN}
    \end{subfigure}
    \begin{subfigure}[tb]{0.46\linewidth}
        \centering
        \includegraphics[width=\linewidth]{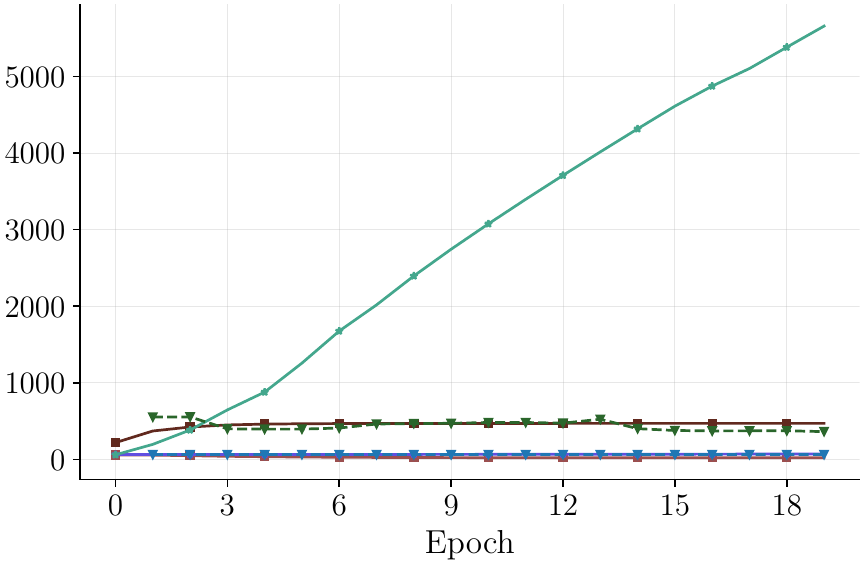}
        \caption{ResNet34}
    \end{subfigure}
    \caption{Frobenius norm of ECAPA-TDNN and ResNet34 computed at different epochs. \texttt{AdaBreg} shows a clear tendency to learn larger weights, as shown by both the adaptive and non-adaptive optimizers. At epoch number $8$ in ECAPA-TDNN and $7$ in ResNet34, the sparsity of non-adaptive \texttt{AdaBreg} is exactly at $90\%$ (not visible from the figure). Despite possessing the exact sparsity rates, the Frobenius norm recorded by the adaptive \texttt{AdaBreg} $90\%$ is much smaller than its non-adaptive counterparts.}
    \label{fig:l2_norm_coparison}
\end{figure}


\subsection{Analyzing the network's norm}
Figure \ref{fig:l2_norm_coparison} shows the Frobenius norm for both \texttt{AdaBreg} and \texttt{LinBreg} trained models in the adaptive and non-adaptive schemes and their dense counterparts.
The trends clearly show that \texttt{AdaBreg} is biased towards learning networks with larger weights relative to \texttt{LinBreg}. The same trend also applies when comparing  \texttt{AdamW} with \texttt{SGD}, albeit with less pronounced differences. This seems to indicate that the Adam-style momenta are the root cause of this phenomenon.

Interestingly, the adaptive versus non-adaptive \texttt{AdaBreg} show a common trend; the network's norm increases as training continues. However, the rate of increase is roughly linear in the non-adaptive case and sub-linear in the adaptive one. It is likely that $\lambda$ adaptation (typically an increase) throughout training in the adaptive case is what controls the rate of increase in the norm of the network. In the non-adaptive case, we observed that $\mathsf{s}(\theta)$ decreases as training continues, which explains the continuous increase in the norm.

Although a smaller norm is associated with higher stability, we found that \texttt{AdaBreg} typically outperforms \texttt{LinBreg}, including on \ac{OOD} data, as discussed in Section~\ref{sec:main_results}.

\begin{figure}[tb]
    \centering
    \begin{subfigure}[tb]{\textwidth}
    \centering
      \includegraphics[width=0.99\linewidth]{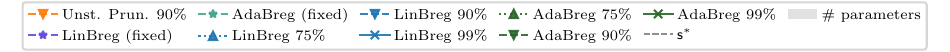}  
    \end{subfigure}
    \begin{subfigure}[tb]{0.5\textwidth}
    \centering
      \includegraphics[width=\linewidth]{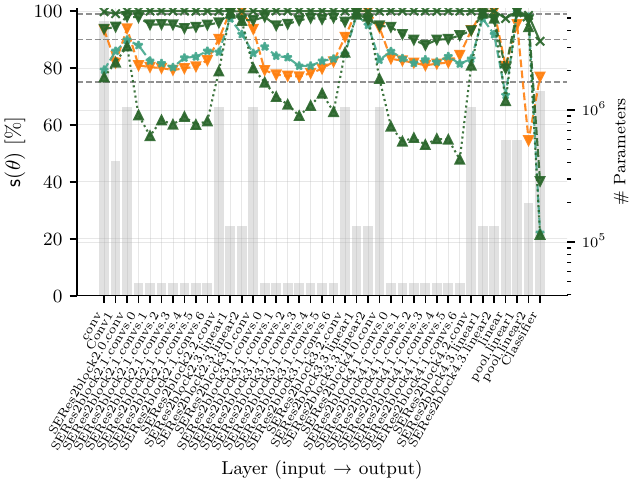} 
      \caption{\texttt{AdaBreg}}
    \end{subfigure}  
    \begin{subfigure}[tb]{0.49\textwidth}
    \centering
      \includegraphics[width=\linewidth]{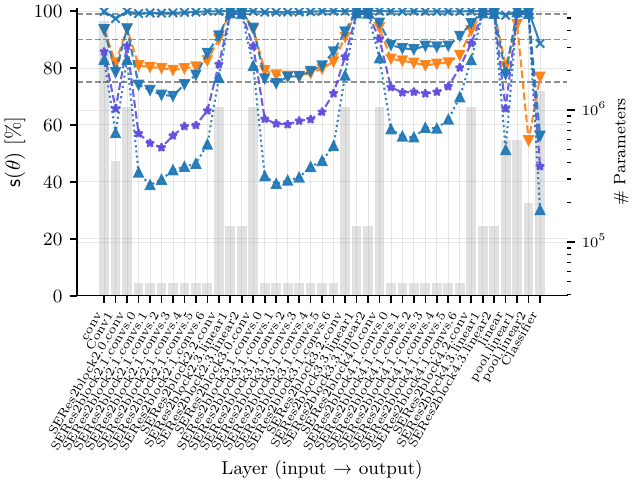} 
      \caption{\texttt{LinBreg}}
    \end{subfigure}    
    \caption{Layer-wise sparsity distribution for Bregman optimizers versus pruning with a gradual schedule. All methods exhibit the same tendency to keep the classifier more dense than the target $\mathsf{s}^\ast$, however, it is more pronounced for \texttt{AdaBreg}. Also, the adaptation of $\lambda$ appears to maintain the sparsity patterns obtained by their non-adaptive counterparts.}
    \label{fig:sparsity_patterns_fixed}
\end{figure}

\subsection{Learned sparsity patterns}
In Figure~\ref{fig:sparsity_patterns_fixed}, we show the sparsity assignment per layer for each ECAPA-TDNN model. Both optimizers have the same tendency to allocate more sparsity budget for the classifier. This is especially pronounced in the case of \texttt{AdaBreg}. 
Since the classifier accounts for a larger fraction of the network in ResNet34 than in ECAPA-TDNN, an even larger share of its weights are driven to high sparsity in order to meet the target $\mathsf{s}^\ast$. 
An obvious remedy is to enforce a minimum sparsity on certain layers or to control the classifier. Finding an optimal sparsity allocation strategy for an arbitrary model is an interesting future endeavor, but it is outside the scope of this work.

Another crucial aspect is the relationship between the sparsity profiles generated by the adaptive and non-adaptive optimizers. Although a one-to-one comparison is challenging due to differences in the sparsity, the distribution patterns exhibit notable similarities across all network layers.

\begin{figure}[tb]
    \centering
    \includegraphics[width=0.65\linewidth]{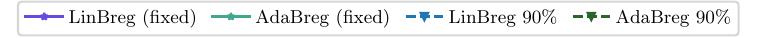}
    \includegraphics[width=\linewidth]{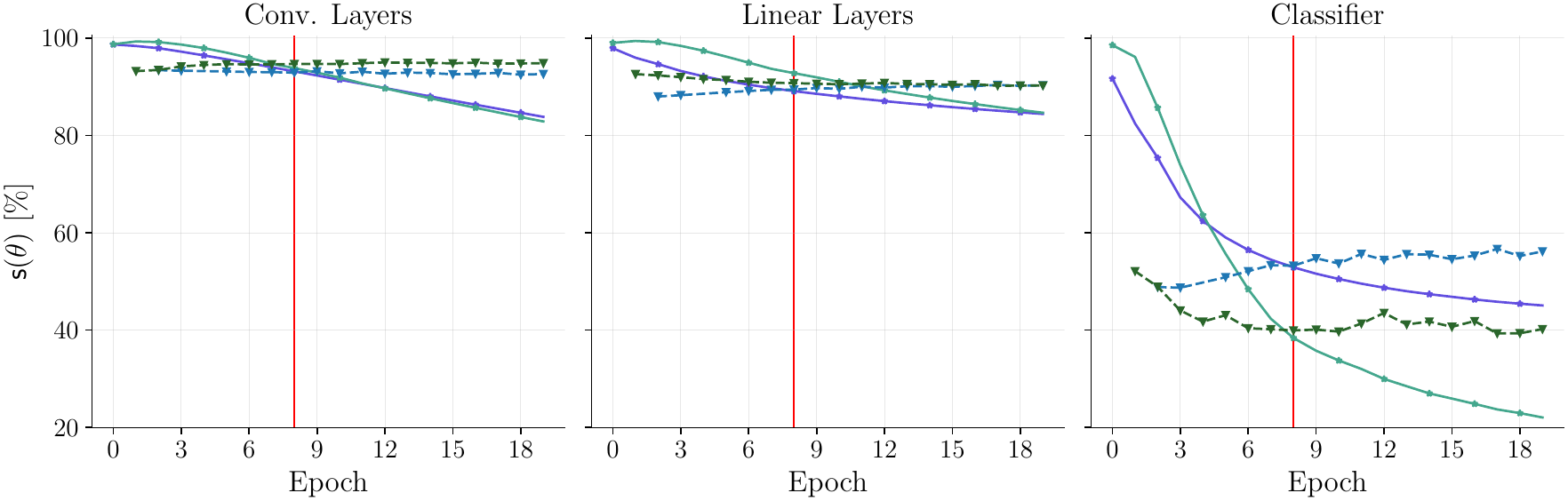}
    \caption{Sparsity of adaptive and non-adaptive Bregman optimizer throughout training computed over different layer groups. At epoch $8$ (highlighted in red), all optimizers achieve a global sparsity of $90\%$. The breakdown of sparsity per layer group suggests that when the global sparsity is the same, adaptive and non-adaptive optimizers learn similar underlying architectures.}
    \label{fig:layer_groups}
\end{figure}

To establish a more precise comparison between \texttt{AdaBreg} and \texttt{LinBreg}, we focus the analysis on epoch $8$ in Figure~\ref{fig:layer_groups}, examining the sparsity distribution across layer groups. At this specific epoch, both optimizers and methods yield nearly identical global sparsity of $90\%$. Under that condition, sparsity assignments per layer group between the adaptive and non-adaptive schemes show a striking similarity. This alignment suggests that, when evaluated at the same sparsity, the underlying architectures learned by the adaptive and non-adaptive methods strongly resemble one another.

\subsection{Ablation study}
As discussed in Section~\ref{sec:sparsity_patterns}, Bregman optimizers leave the classifier considerably denser than the rest of the network. This is problematic for two reasons: (1) the classifier absorbs the tightly constrained global sparsity budget, forcing intermediate layers into near-complete sparsity; and (2) this budget is ultimately "wasted", as the classifier itself is discarded during inference. As shown in Figure~\ref{fig:sparsity_patterns_fixed}, this behavior persists in both \texttt{AdaBreg} and \texttt{LinBreg}, with and without $\lambda$ adaptation.

Based on this reasoning, it is expected that penalizing the classifier more heavily should redistribute the sparsity budget toward the intermediate layers and therefore improve performance.
To verify this, we apply a simple per-group rescaling such that $\lambda \leftarrow a\lambda$ ($a{=}2$ for the classifier, $a{=}1$ otherwise). Table~\ref{tab:before_and_after_2lambda} demonstrates that at $\mathsf{s}^\ast{=}99\%$, this adjustment consistently improves performance across both optimizers and test sets. The most notable improvement is the performance recovery of the \texttt{AdaBreg}-trained ResNet34 on \voxceleb. Prior to the added penalty, it showed near-random performance (roughly 50\% \ac{EER}). After adding the classifier's penalty, the \ac{EER} jumps to, e.g., $7.97\%$ on Vox1-O, significantly closing the gap with ECAPA-TDNN.

\begin{table}[tb]
    \centering
    \footnotesize
    \caption{EER at $99\%$ sparsity before and after applying the classifier penalty across evaluation datasets. "+ cls. $2\lambda$" refers to our method and using $2\lambda$ in proximal threshold for the classifier but $\lambda$ for the rest of the network.}
    \label{tab:before_and_after_2lambda}
    \renewcommand{\arraystretch}{0.8}
    \setlength{\tabcolsep}{4pt}
    \begin{tabular}{@{} ll *{8}{S[table-format=2.2]} @{}}
    \toprule
    \multirow{2}{*}{\textbf{Optimizer}} & \multirow{2}{*}{\textbf{Architecture}} & \multicolumn{2}{c}{\textbf{Vox1-O}} & \multicolumn{2}{c}{\textbf{Vox1-E}} & \multicolumn{2}{c}{\textbf{Vox1-H}} & \multicolumn{2}{c}{\textbf{CNCeleb-E}} \\
    \cmidrule(lr){3-4} \cmidrule(lr){5-6} \cmidrule(lr){7-8} \cmidrule(l){9-10}
    & & \textbf{Proposed} & \textbf{+ cls. $2\lambda$} & \textbf{Proposed} & \textbf{+ cls. $2\lambda$} & \textbf{Proposed} & \textbf{+ cls. $2\lambda$} & \textbf{Proposed} & \textbf{+ cls. $2\lambda$} \\ 
    \midrule
    \multirow{2}{*}{\textbf{AdaBreg}} 
    & ECAPA-TDNN  & 7.25 & \graycell{6.05}  & 7.45 & \graycell{6.36} & 10.44 & \graycell{10.06} & 19.00 & \graycell{18.84} \\
    & ResNet34    & 49.43 & \graycell{7.97} & 49.06 & \graycell{7.81} & 49.28 & \graycell{11.35} & 45.96 & \graycell{17.26} \\ 
    \midrule
    \multirow{2}{*}{\textbf{LinBreg}} 
    & ECAPA-TDNN  & 6.13 & \graycell{5.56} & 6.44 & \graycell{5.59} & 9.28 & \graycell{8.21} & 16.27 & \graycell{15.63} \\
    & ResNet34    & 7.13 & \graycell{5.94} & 7.34 & \graycell{6.34} & 10.70 & \graycell{9.35} & 16.62 & \graycell{15.70} \\ 
    \bottomrule
    \end{tabular}
\end{table}

\subsection{Results for prox rescaling and subgradient correction}\label{sec:subgradcorr_proxrescale}                                                                   
We report convergence results for ECAPA-TDNN on \cnceleb for the settings introduced in Appendices~\ref{app:proxrescale} and~\ref{app:corr}.

For the subgradient correction, \texttt{LinBreg} demonstrates similar behavior to the proposed adaptation for sparsity rates of $90\%$ and $95\%$. At $99\%$ sparsity, the proposed adaptation reaches higher accuracies, but the subgradient-corrected variant is notably more stable. For \texttt{AdaBreg}, our adaptation achieves higher training and validation accuracies. An investigation of the sparsity and $\lambda$ profiles clearly indicates that the values of $\lambda$ increase dramatically for \texttt{AdaBreg} and only cap at $10^{3}$, which is an implemented safeguard. Despite the sharp increase, the sparsity profiles remain well-behaved at high $\lambda$ values; the cap only prevents the models from reaching the higher sparsity targets at $95\%$ and $99\%$.

\begin{figure}[tb]
    \centering
    \includegraphics[width=0.8\linewidth]{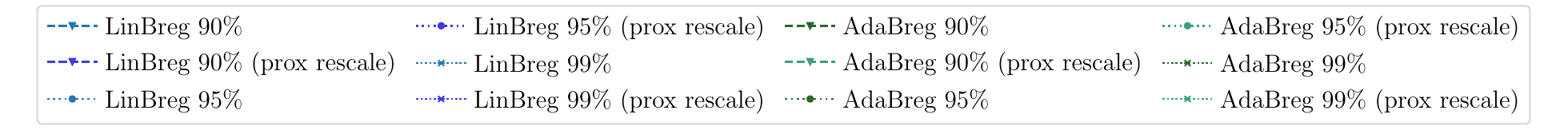}
    \includegraphics[width=0.5\linewidth]{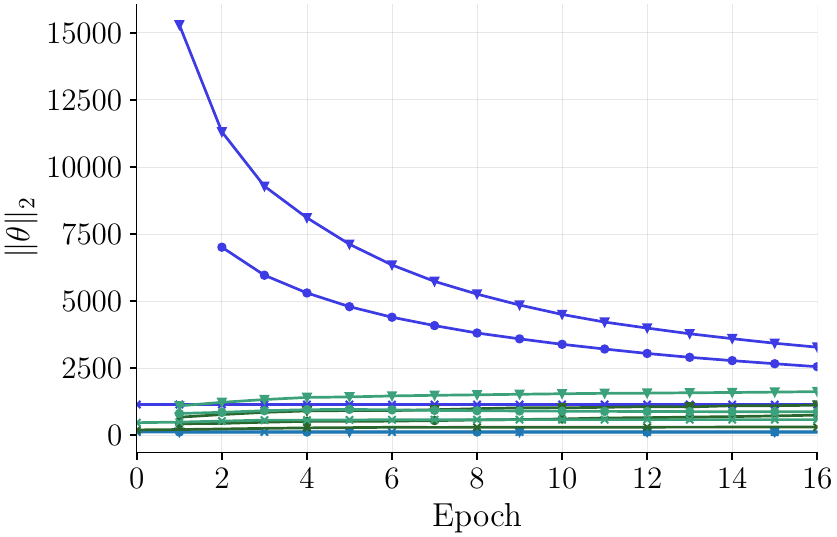}
    \caption{Frobenius norm of ECAPA-TDNN. In the $\operatorname{prox}$-rescaling scheme, \texttt{LinBreg}'s norm explodes due to dividing weights by small $\lambda$, whereas \texttt{AdaBreg}'s norm decreases because $\lambda$ typically adapts to values $> 1$.}
    \label{fig:forb_norm_prox_rescale}
\end{figure}

For the prox-rescaled variants, \texttt{AdaBreg} shows on-par convergence to our proposed method at $90\%$ and $95\%$ sparsity. At $99\%$, the relationship is harder to characterize since accuracies fluctuate intensely. However, when considering \texttt{LinBreg}, the proposed adaptation is clearly better. This is likely due to adapting to values of $\lambda$ that are $\ll 1$, resulting in increased weight magnitudes, as depicted in Figure~\ref{fig:forb_norm_prox_rescale}.

\begin{figure}[tb]
    \centering
    \includegraphics[width=0.8\linewidth]{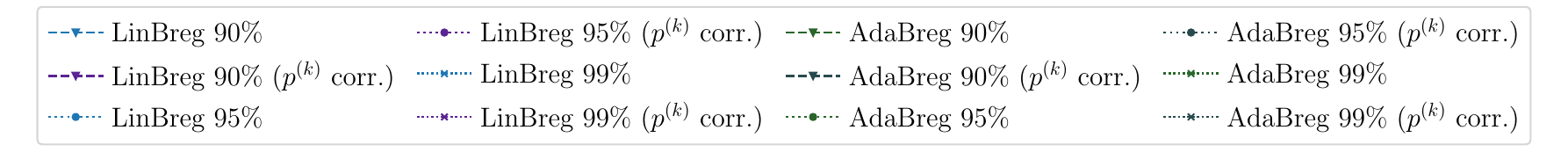}
    \includegraphics[width=0.5\linewidth]{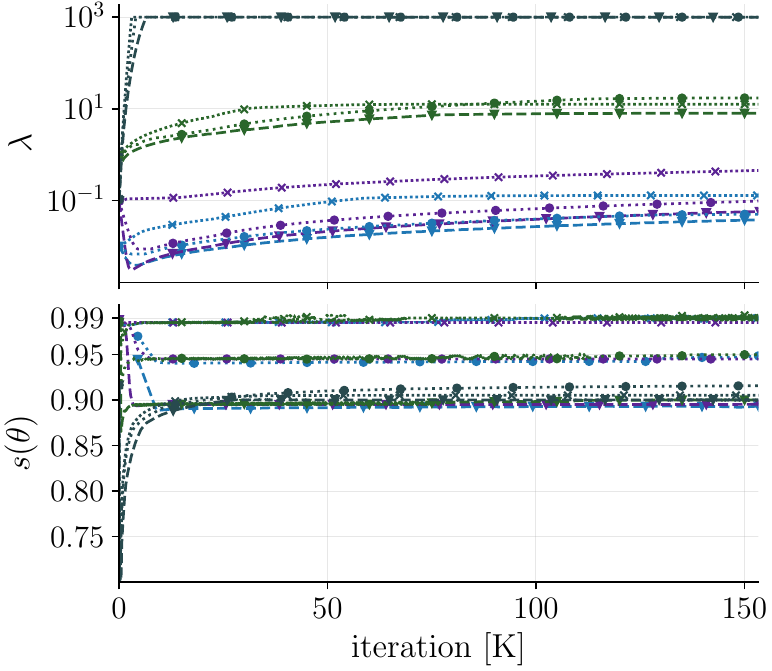}  
    \caption{$\lambda$ and sparsity profiles of the proposed adaptation versus the subgradient-corrected variant on \cnceleb. With subgradient correction, \texttt{LinBreg} achieves the target sparsity at slightly higher $\lambda$  values. For \texttt{AdaBreg}, $\lambda$ increases dramatically and caps at $10^{3}$ (an implemented safeguard). Although sparsity remains well-behaved at high $\lambda$ values, the cap prevents the models from reaching the higher sparsity targets at $95\%$ and $99\%$.}
    \label{fig:subgrad_corr_internal_curves}
\end{figure}
\begin{figure}[tb]
    \centering
    \includegraphics[width=0.8\linewidth]{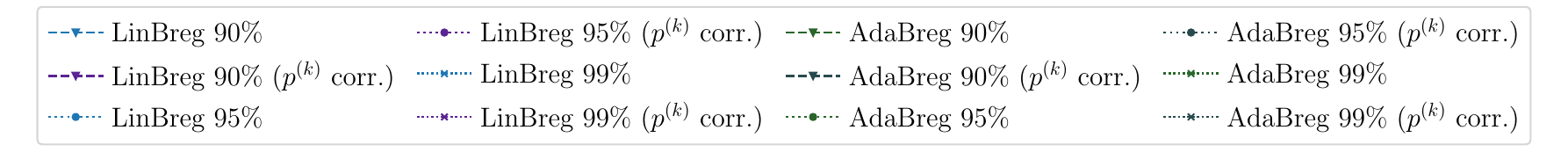}\\
    \begin{subfigure}[tb]{0.49\textwidth}
    \includegraphics[width=\linewidth]{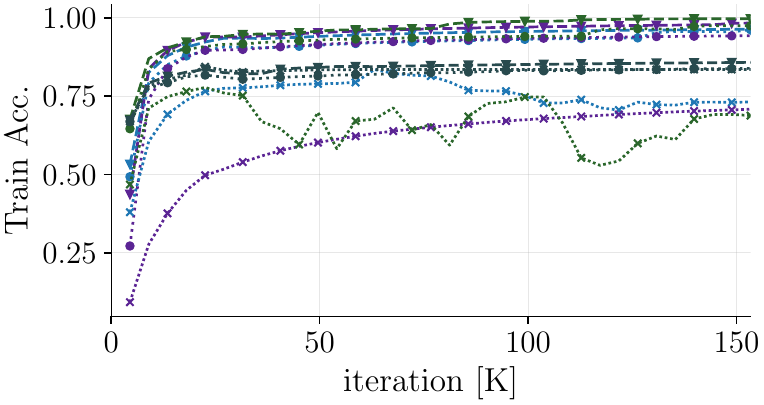}          
    \end{subfigure}
    \begin{subfigure}[tb]{0.49\textwidth}
    \includegraphics[width=\linewidth]{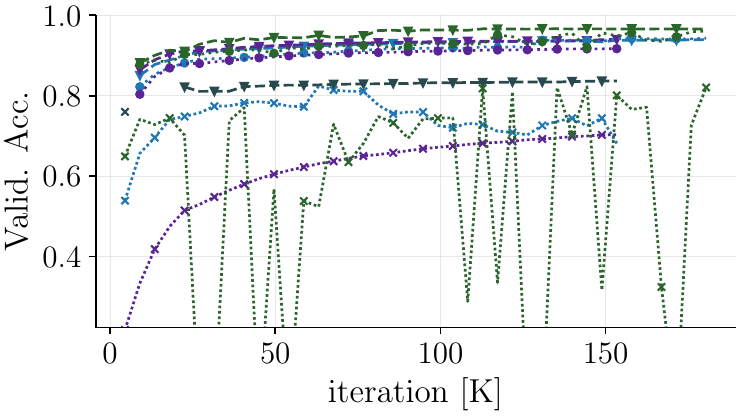}          
    \end{subfigure}
    \caption{Convergence of the subgradient-corrected variant compared to the proposed adaptation for ECAPA-TDNN on \cnceleb.}
    \label{fig:subgrad_convergence}
\end{figure}
\begin{figure}[tb]
    \centering
    \includegraphics[width=0.8\linewidth]{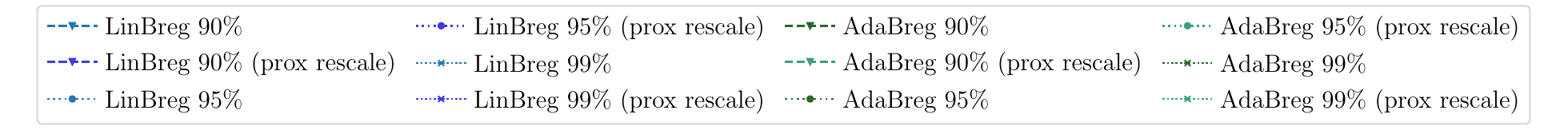}\\
    \begin{subfigure}[tb]{0.49\textwidth}
    \includegraphics[width=\linewidth]{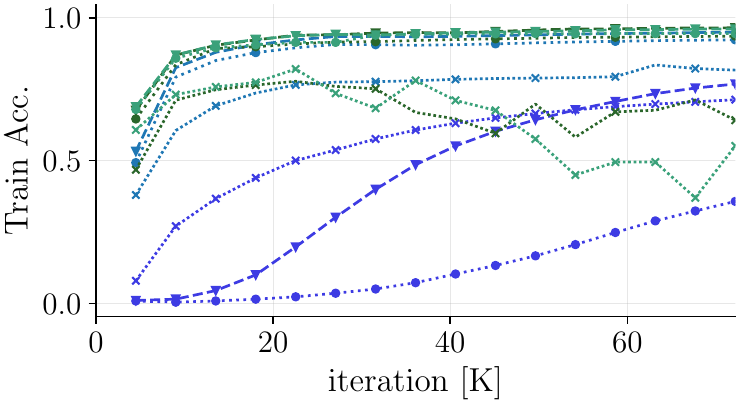}          
    \end{subfigure}
    \begin{subfigure}[tb]{0.49\textwidth}
    \includegraphics[width=\linewidth]{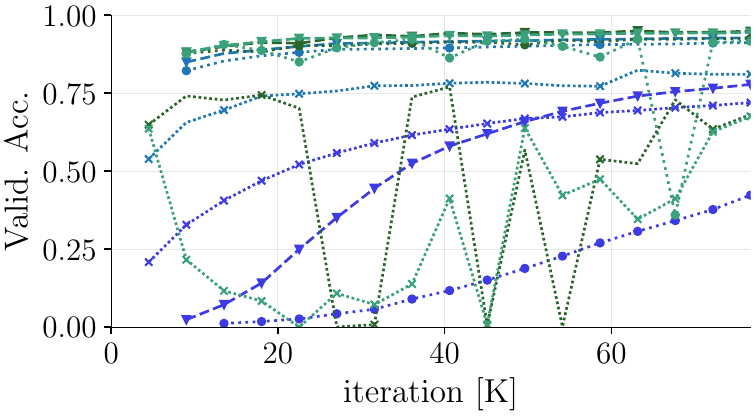}          
    \end{subfigure}
    \caption{Convergence of the $\operatorname{prox}$-rescaled variant compared to the proposed adaptation for ECAPA-TDNN on \cnceleb.}
    \label{fig:prox_rescale_convergence}
\end{figure}

\clearpage

\end{document}